\documentclass{article}
\usepackage{lmodern}
\usepackage{microtype}
\usepackage{graphicx}
\usepackage{subcaption}
\usepackage{booktabs} 
\usepackage{multirow}
\usepackage{amsmath}

\usepackage{booktabs}
\usepackage{siunitx}
\usepackage{colortbl}
\usepackage[most]{tcolorbox}

\usepackage{graphicx}
\usepackage{subcaption}
\usepackage{booktabs}
\usepackage{multirow}
\usepackage{graphicx}
\usepackage[table,xcdraw]{xcolor} 
\definecolor{bestblue}{RGB}{200, 230, 255}  
\definecolor{secondblue}{RGB}{235, 245, 255} 

\usepackage{multirow}   
\usepackage{graphicx}   
\usepackage[table]{xcolor} 

\definecolor{designblue}{HTML}{1976D2}

\newcommand{\designtitle}[2]{%
  \par\medskip\noindent
  {%
    \begingroup
    \setlength{\fboxsep}{6pt}
    \colorbox{#1!10}{
      \strut
      \makebox[\dimexpr\linewidth-2\fboxsep\relax][l]{%
        \textcolor{#1}{\rule{2.5pt}{1.6ex}}%
        \hspace{0.8em}%
        \textbf{#2}%
      }%
    }%
    \endgroup
  }%
  \par\smallskip
}
\definecolor{bestgreen}{HTML}{C6E3B4}
\definecolor{secondyellow}{HTML}{FFF2B2}

\usepackage{hyperref}

\usepackage[preprint]{icml2026}

\usepackage{amsmath}
\usepackage{amssymb}
\usepackage{mathtools}
\usepackage{amsthm}

\usepackage[capitalize,noabbrev]{cleveref}

\theoremstyle{plain}

\theoremstyle{definition}

\theoremstyle{remark}

\usepackage[textsize=tiny]{todonotes}

\icmltitlerunning{\textit{\textbf{UGID}}: Unified Graph Isomorphism for Debiasing Large Language Models}

\begin{document}

\twocolumn[
  \icmltitle{\textit{\textbf{UGID}}: Unified Graph Isomorphism for Debiasing Large Language Models}



  \icmlsetsymbol{equal}{*}

  \begin{icmlauthorlist}
    \icmlauthor{Zikang Ding}{yyy,comp}
    \icmlauthor{Junchi Yao}{yyy,comp}
    \icmlauthor{Junhao Li}{sch}
    \icmlauthor{Yi Zhang}{sch}
    \icmlauthor{Wenbo Jiang}{yyy}
    \icmlauthor{Hongbo Liu}{yyy}
    \icmlauthor{Lijie Hu}{comp}
  \end{icmlauthorlist}

  \icmlaffiliation{yyy}{University of Electronic Science and Technology of China, Chengdu, China}
  \icmlaffiliation{comp}{Mohamed bin Zayed University of Artificial Intelligence, Abu Dhabi, United Arab Emirates}
  \icmlaffiliation{sch}{South China University of Technology, Guangzhou, China}

  \icmlcorrespondingauthor{Lijie Hu}{lijie.hu@mbzuai.ac.ae}

  \icmlkeywords{Machine Learning, ICML}

  \vskip 0.3in
]



\printAffiliationsAndNotice{}  

\begin{abstract}
Large language models (LLMs) exhibit pronounced social biases. Output-level or data-optimization--based debiasing methods cannot fully resolve these biases, and many prior works have shown that biases are embedded in internal representations. We propose \underline{U}nified \underline{G}raph \underline{I}somorphism for \underline{D}ebiasing large language models (\textit{\textbf{UGID}}), an internal-representation--level debiasing framework for large language models that models the Transformer as a structured computational graph, where attention mechanisms define the routing edges of the graph and hidden states define the graph nodes. Specifically, debiasing is formulated as enforcing invariance of the graph structure across counterfactual inputs, with differences allowed only on sensitive attributes. \textit{\textbf{UGID}} jointly constrains attention routing and hidden representations in bias-sensitive regions, effectively preventing bias migration across architectural components. To achieve effective behavioral alignment without degrading general capabilities, we introduce a log-space constraint on sensitive logits and a selective anchor-based objective to preserve definitional semantics. Extensive experiments on large language models demonstrate that \textit{\textbf{UGID}} effectively reduces bias under both in-distribution and out-of-distribution settings, significantly reduces internal structural discrepancies, and preserves model safety and utility.

\end{abstract}

\section{Introduction}
\begin{figure}[t]
    \centering
    \begin{subfigure}[t]{0.48\textwidth}
        \centering
        \includegraphics[width=\textwidth]
        {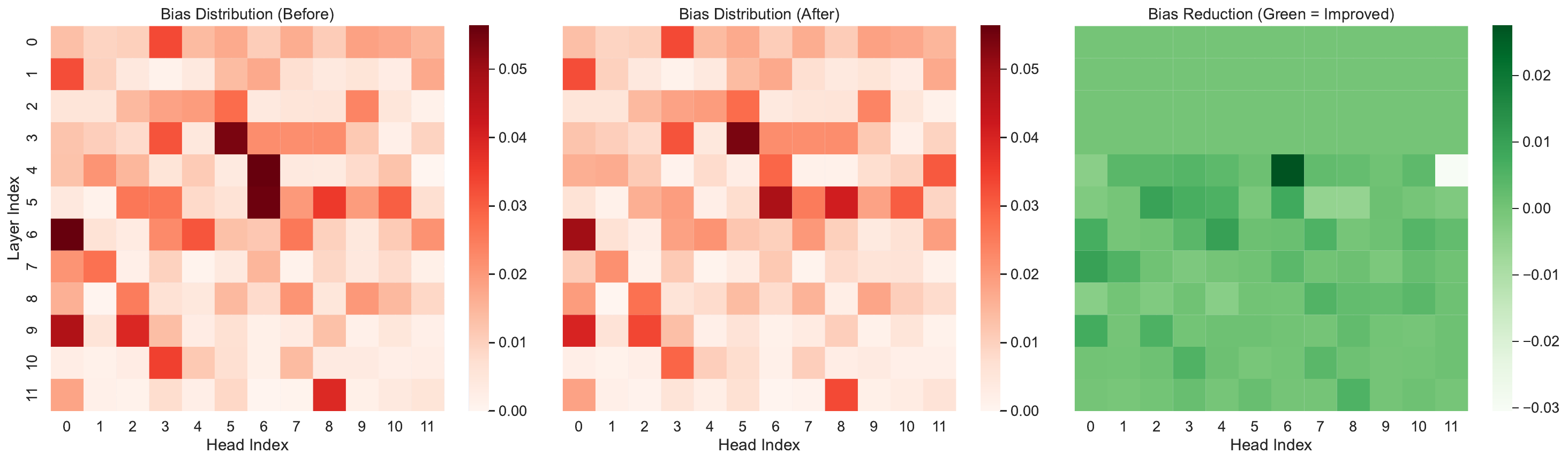}
        \caption{\textbf{GPT-2.} Bias distribution shifts.}
    \end{subfigure}
    
    \vspace{-0.5mm} 

    \begin{subfigure}[t]{0.48\textwidth}
        \centering
        \includegraphics[width=\textwidth]{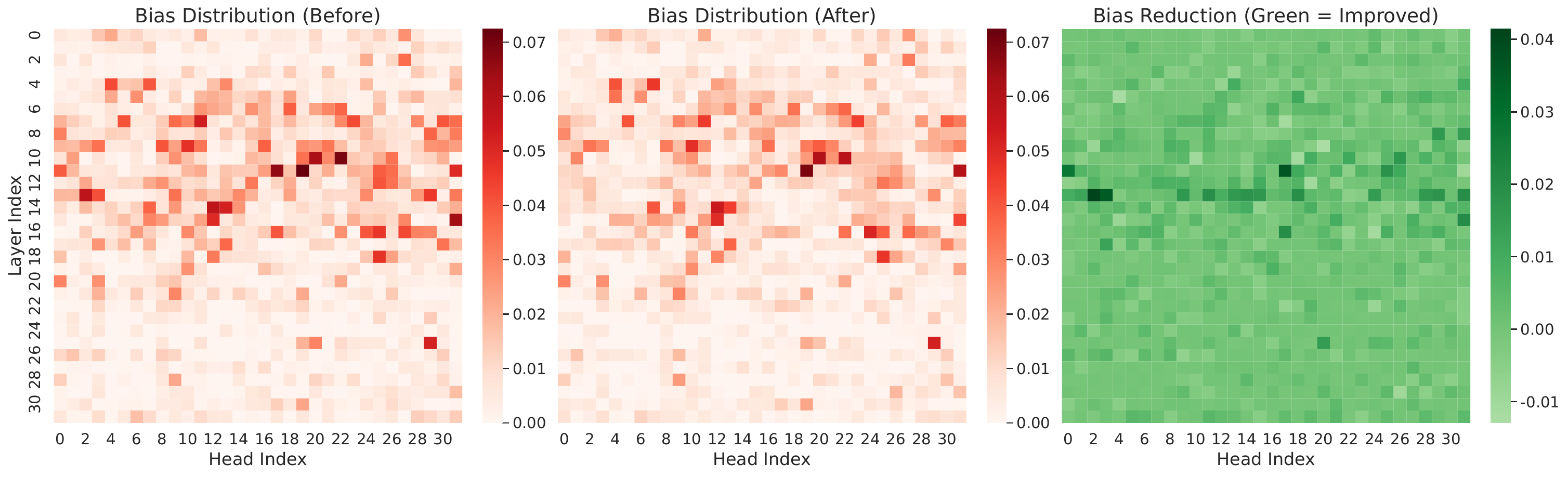}
        \caption{\textbf{LLaMA-3-8B.} Bias distribution shifts.}
    \end{subfigure}

    \caption{Visualization of bias distribution shifts. The vertical layout allows for clear layer-wise comparison between GPT-2 (top) and LLaMA-3 (bottom).}
    \vspace{-4mm}
    \label{fig:bias_map_vertical}
\end{figure}
Large Language Models (LLMs) have become foundational infrastructure for modern artificial intelligence systems, yet they often reflect and amplify social biases latent in the training data \citep{zhao2023survey, huang2023towards, gallegos2024bias, navigli2023biases}. These biases are not merely surface-level artifacts but are deeply encoded within internal representations and computational processes \citep{chandna2025dissecting, cheng2023marked}. Existing debiasing work can be divided into (i) \textit{model-external methods}, which adjust data, inference, or output behavior but do not constrain internal bias representations \citep{zmigrod2019counterfactual, li2024data, kiashemshaki2025simulating, kim2025klaad, gallegos2025self, chandna2025dissecting}; and (ii) \textit{model-internal methods}, which primarily focus on locating or understanding local components like attention heads or FFN vectors \citep{meng2022locating, geva2021transformer, prakash2024interpreting, zhou2024unibias}. However, these existing methods are still difficult to fundamentally weaken the biases present inside the model \citep{fang2025trustworthy, vijjini2025exploring}.

Our investigation reveals the scale-dependence of bias within the internal mechanisms of the model. Through controlled experiments, we observe that when constraints are imposed only on the attention mechanism, bias metrics can be significantly reduced in models with small parameter scales like GPT-2, but this effect is significantly weakened or even fails when the model scale is extended to the billion-parameter level (Figure \ref{fig:bias_map_vertical}). This indicates that bias is not determined solely by attention; instead, feed-forward networks (FFNs), serving as primary storage units for semantic and factual knowledge, also encode and amplify bias \citep{NEURIPS2020_92650b2e, meng2022locating}. Thus, constraining attention routing alone is insufficient to fundamentally weaken the bias embedded within the model.

To overcome these limitations, we propose \textit{\textbf{UGID}}, the first unified distance-invariant debiasing framework that reformulates the problem as a computational graph isomorphism challenge. Leveraging the perspective of mechanistic interpretability, we model the internal bias migration as a dynamic computational graph $\mathcal{G}=(\mathcal{V}, \mathcal{E})$ \citep{binkowski2025hallucination}, where token representations constitute the node set $\mathcal{V}$ \citep{geva2021transformer, meng2022locating} and attention mechanisms define the weighted dependency edge set $\mathcal{E}$ \citep{cai2024locating}. Specifically, we develop a dual alignment framework to achieve graph isomorphism: for edges, we employ a combinatorial Laplacian operator with attention-sink masking to align the spectral characteristics of semantic routing; for nodes, we propose a selective representation alignment strategy to prevent FFNs from bias compensation. Additionally, a log-space behavioral guidance is introduced to stabilize the model's distribution over non-sensitive semantics. Extensive experiments across multiple scales demonstrate that \textit{\textbf{UGID}} achieves state-of-the-art (SOTA) performance in both debiasing effectiveness and structural stability. Mechanistic analyses further verify that our framework effectively decouples bias from internal reasoning topology.

The contributions of our work are summarized as follows:
\begin{itemize}
    \item We propose \textit{\textbf{UGID}}, the first framework to achieve debiasing via Unified Graph Isomorphism. By employing Laplacian Spectral Constraints to precisely align the routing topology (edges) and Selective Anchoring to stabilize semantic memory (nodes), \textit{\textbf{UGID}} mechanistically severs the migration pathways of bias.
    \item We provide mechanistic interpretability evidence for the debiasing process. Through Spectral Diagnostics, Logit Lens, and Activation Patching, we visualize and verify the elimination of bias at a topological level, offering a physical-level explanation for the internal bias governance in LLMs.
    \item Extensive experiments on public datasets demonstrate that \textit{\textbf{UGID}} achieves a better trade-off between debiasing and model performance preservation, consistently outperforming state-of-the-art methods across various model sizes and data scenarios.
\end{itemize}

\begin{figure*}[t]
    \centering
    \includegraphics[width=1\textwidth]{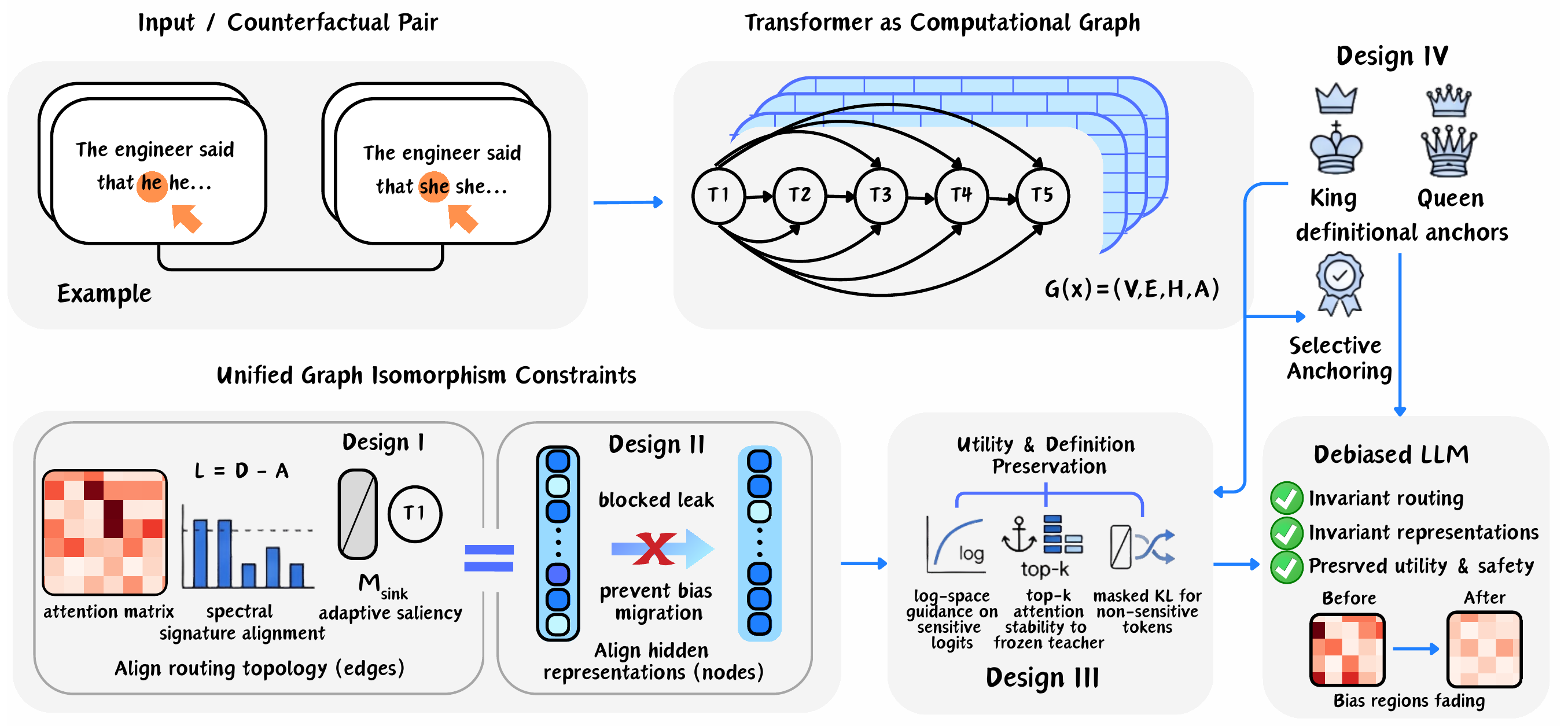} 
    \caption{
        \textbf{Overview of the UGID framework.} 
    }
    \vspace{-3mm}
    \label{fig:framework}
\end{figure*}

\section{Related Work}
\textbf{Debiasing and Performance.} 
Social bias in large language models (LLMs) is not only reflected in their outputs, but is also embedded in the models' internal representations and computational processes. These biases are complex and diverse, ranging from explicit stereotypes captured by benchmarks such as BBQ \citep{parrish2022bbq} and CrowS-Pairs \citep{nangia2020crows}, to more subtle affective disparities observed in datasets like BOLD \citep{dhamala2021bold} and RealToxicityPrompts \citep{gehman2020realtoxicityprompts}. As a result, balancing effective debiasing with preserving model performance remains a fundamental challenge. Existing debiasing methods often face a bottleneck: stronger debiasing capability can come at the cost of the model's original utility \citep{cheng2023marked}. For example, data-optimization-driven approaches such as CDA \citep{zmigrod2019counterfactual} may reduce bias while degrading the model's intrinsic performance \citep{kiashemshaki2025simulating}. In contrast, external methods such as Self-Debias \citep{gallegos2025self}, KLAAD \citep{kim2025klaad}, and BiasFilter \citep{cheng2025biasfilter} essentially intervene at the behavioral level. These approaches also struggle to maintain general reasoning ability and are vulnerable to prompt rephrasing or out-of-distribution (OOD) scenarios, because they do not fundamentally weaken the internal bias structures of the model.

\textbf{Mechanistic Analysis and Structural Interventions.} 
Mechanistic interpretability offers a clearer lens to investigate these internal pathways, identifying bias-carrying attention heads via causal mediation \citep{vig2020investigating, cai2024locating} and biased "Key-Value Memories" in Feed-Forward Networks (FFNs) \citep{geva2021transformer, meng2022locating}. Crucially, our research reveals that bias is scale-dependent: interventions on attention alone may effectively reduce bias in small-scale models like GPT-2 but fail in billion-parameter models like LLaMA-3-8B due to \textit{bias migration} into FFN components. This phenomenon suggests that a robust debiasing framework must jointly address attention routing and hidden representations to achieve structural invariance. 

While recent research has explored graph-structured removal \citep{arduini2020adversarial}, specific bias subspaces \citep{yifei2023conceptor}, and causal adjustments \citep{wu2025beyond, zhang2024causal}, these methods typically focus on isolated components and fail to arrest the holistic migration of bias across architectural layers. Furthermore, although \citet{liu2024lidao} advocates for limited interventions to preserve utility, and other studies leverage structured knowledge or internal representations \citep{ma2024debiasing, zhang2024grace, wang2025graph}, they do not enforce the topological isomorphism necessary to handle out-of-distribution (OOD) scenarios or complex bias migration pathways. UGID addresses these gaps by reformulating debiasing as a unified structural alignment problem, ensuring that the reasoning topology remains invariant across counterfactual inputs and effectively severing bias migration pathways throughout the computational graph.

Overall, existing work provides important insights into bias localization and partial structural interventions, but does not offer a unified constraint over the global computation structure under counterfactual perturbations. 
UGID is motivated by this gap and explores debiasing from a structural alignment perspective.

\section{Methodology}
\label{sec:method}
This section introduces \textit{\textbf{UGID}}, a unified framework for structural bias correction. The overall architecture and workflow of our framework are illustrated in Figure \ref{fig:framework}. First, we formally define the problem. Then, we present detailed solutions based on the following four key questions: (I) How to align the routing (Edges)?, (II) RHow to prevent bias migration (Nodes)?, (III) How to preserve model utility?, and (IV) How to handle definitional bias?

\subsection{Problem Formulation}

We model the Transformer layer $l$ as a dynamic computational graph $\mathcal{G}_l(x) = (\mathcal{V}, \mathcal{E}, \mathbf{H}_l, \mathbf{A}_l)$, where nodes $\mathcal{V}$ represent tokens, node features $\mathbf{H}_l \in \mathbb{R}^{T \times d}$ represent hidden states, and weighted edges $\mathbf{A}_l \in \mathbb{R}^{H \times T \times T}$ represent attention mechanisms.

Given a counterfactual pair $(x, x')$ differing only in a sensitive attribute (e.g., \textit{he} vs. \textit{she}), our goal is to enforce \textit{Unified Graph Isomorphism} in bias-sensitive regions. That is, the reasoning topology and semantic representations should remain invariant:
{\small
\begin{equation}
    \mathcal{G}_l(x) \cong \mathcal{G}_l(x') \iff \underbrace{\mathbf{A}_l(x) \approx \mathbf{A}_l(x')}_{\text{Design I}} \land \underbrace{\mathbf{H}_l(x) \approx \mathbf{H}_l(x')}_{\text{Design II}},
\end{equation}}

where $\mathcal{S}_{\text{target}}$ denotes the set of target layers
at which structural constraints are applied.

\subsection{UGID}
\designtitle{designblue}{Design I: How to align the routing (Edges)?}
One of the main challenges in debiasing is routing bias, where attention heads form a fixed bottleneck around sensitive tokens. To address this issue, we propose an adaptive spectral-invariant training method. Rather than matching the raw attention weights, we align the spectral features of the attention graphs. For a target layer $l$, we follow \citet{binkowski2025hallucination} to construct a combined Laplacian matrix $\mathbf{L} = \mathbf{D} - \mathbf{A}$ to capture the topological structure of the attention graph. Meanwhile, we use the normalized in-degree matrix $\mathbf{D}$ to account for varying context lengths.

Due to the causal nature of autoregressive LLMs, the attention matrix $\mathbf{A}$ is strictly lower-triangular. we compute the eigenvalues $\boldsymbol{\lambda}$ directly from the diagonal elements without expensive eigendecomposition: $\lambda_t = d_{tt} - A_{tt}$. This reduces the complexity to $\mathcal{O}(T)$, making training-time regularization computationally feasible. However, standard spectral matching is compromised by the ``Attention Sink'' phenomenon. As formally identified by \citet{xiao2023efficient} and rigorously characterized in recent work by \citet{qiu2025gated}, this phenomenon is an inherent artifact of the Softmax normalization, where the initial token absorbs excess attention mass to satisfy probability constraints. To address this, we introduce an Attention Sink Mask($\mathbf{M}^{sink}$) to filter out the first token, and an Adaptive Saliency Weight ($\omega$) derived from the pronoun's attention to focus on bias hubs.

The Attention Sink Mask $\mathbf{M}^{\text{sink}} \in \{0,1\}^{T}$ excludes the first token position.
Here, $p$ denotes the token index corresponding to the sensitive pronoun
(e.g., \textit{he} or \textit{she}). The edge loss is defined as:
{\small
\begin{align}
\mathcal{L}_{\text{edge}}
&= \sum_{l \in \mathcal{S}_{\text{target}}}
\frac{1}{H \sum \mathbf{M}^{\text{sink}}}
\sum_{h=1}^{H} \sum_{t=1}^{T}
\mathbf{M}^{\text{sink}}_t \,
\text{sg}(\omega_{l,h,t}) \notag \\
&\qquad \cdot
\big( \lambda_{l,h,t}(x) - \lambda_{l,h,t}(x') \big)^2,
\end{align}}
where $\omega_{l,h,t} = \frac{1}{2} (\mathbf{A}_{l,h,p,t}(x) + \mathbf{A}_{l,h,p,t}(x'))$, and $\text{sg}(\cdot)$ is the stop-gradient operator. This design ensures the reasoning topology remains invariant to gender perturbations.

\designtitle{designblue}{Design II: How to prevent bias migration (Nodes)?}

Under the condition that only attention bias is constrained, bias migrates into the hidden states (FNNs). To address this issue, we must perform node isomorphism. We constrain the hidden representations $\mathbf{H}_l$ to remain consistent across counterfactual contexts, thereby effectively preventing bias from being stored in the value vectors. Since hidden states are shared across heads, we aggregate the saliency weights:
{\small
\begin{align}
\mathcal{L}_{\text{node}}
&= \sum_{l \in \mathcal{S}_{\text{target}}}
\frac{1}{\sum \mathbf{M}^{\text{sink}}}
\sum_{t=1}^{T}
\mathbf{M}^{\text{sink}}_t \,
\bar{\omega}_{l,t} \notag \\
&\qquad \cdot
\left\| \mathbf{H}_{l,t}(x) - \mathbf{H}_{l,t}(x') \right\|_2^2.
\end{align}}
By jointly optimizing Design I and II, we achieve holographic alignment of the computational graph.

\designtitle{designblue}{Design III: How to preserve model utility?}
Aggressive structural updates can lead to catastrophic forgetting. To ensure stability, we introduce a composite stability objective consisting of three terms:

\textbf{1. Log-Space Guidance.} To prevent gradient vanishing on rare tokens, we penalize log-probability divergence on sensitive attributes:
{\small
\begin{equation}
\mathcal{L}_{logit} = \left(\log P_\theta(v_{he}|x) - \log P_\theta(v_{she}|x')\right)^2.
\end{equation}
}
\textbf{2. Top-K Stability.} We anchor the student's attention patterns to the frozen teacher $P_{ref}$ to preserve syntactic correctness:
{\small
\begin{equation}
\mathcal{L}_{topk} = \sum_{l,h} \left\| (\mathbf{A}^\theta - \mathbf{A}^{ref}) \odot \mathbb{I}_{topk}(\mathbf{A}^{ref}) \right\|_1.
\end{equation}
}
\textbf{3. Semantic Preservation.} We apply a symmetric, masked KL divergence to retain general knowledge while allowing changes on sensitive tokens:
{\small
\begin{equation}
\begin{aligned}
\mathcal{L}_{\mathrm{KL}}
&= \frac{1}{2} \sum_{x \in \{x, x'\}} \sum_{t=1}^{T}
(1 - \mathbf{M}^{\mathrm{sens}}_t) \\
&\quad \cdot
D_{\mathrm{KL}}\!\left(
P_\theta(\cdot \mid x_{<t})
\;\middle\|\;
P_{\mathrm{ref}}(\cdot \mid x_{<t})
\right),
\end{aligned}
\end{equation}
}
where, $\mathbf{M}^{\mathrm{sens}} \in \{0,1\}^{T}$ denotes a binary mask indicating positions corresponding to sensitive attributes.
\designtitle{designblue}{Design IV: How to handle definitional bias?}

A critical risk in debiasing is \textit{Concept Erasure}, where the model loses the ability to distinguish gender in definitional contexts (e.g., ``King'' vs. ``Queen''). To address this, we propose a Selective Alignment Strategy.
We construct the dataset $\mathcal{D}$ as a mixture of target pairs (stereotypes) and anchor pairs (definitions).

$\mathcal{D}_{\text{target}}$ contains stereotypical counterfactual pairs,
while $\mathcal{D}_{\text{anchor}}$ consists of definitional anchor pairs. The loss function adapts dynamically:
{\small
\begin{equation}
    \mathcal{L}_{batch} =
    \begin{cases}
    \gamma_{e}\mathcal{L}_{edge} + \gamma_{n}\mathcal{L}_{node} + \mathcal{L}_{aux} & \text{if } x \in \mathcal{D}_{target} \\
    \lambda_{anchor}\cdot \mathcal{L}^{anchor}_{KL} & \text{if } x \in \mathcal{D}_{anchor}
    \end{cases}
\end{equation}
}
For anchor data, we apply standard unmasked KL divergence ($\mathcal{L}^{anchor}_{KL}$) to strictly enforce the preservation of definitional gender semantics.

The complete training procedure is summarized in Appendix~\ref{Algorithm}.

\definecolor{paleblue}{RGB}{235, 245, 255}

\begin{table*}[t]
\centering
\renewcommand{\arraystretch}{1.2}
\setlength{\tabcolsep}{5pt}
\resizebox{\textwidth}{!}{
\begin{tabular}{lccccccccc}
\toprule
\multirow{3}{*}{\textbf{Method}} 
& \multicolumn{4}{c}{\textbf{Debiasing Effectiveness} ($\downarrow$ 1.0)} 
& \multicolumn{2}{c}{\textbf{Mechanism} ($\downarrow$ 0)} 
& \textbf{Safety} 
& \multicolumn{2}{c}{\textbf{Utility}} \\
\cmidrule(lr){2-5} \cmidrule(lr){6-7} \cmidrule(lr){8-8} \cmidrule(lr){9-10}
& \multicolumn{2}{c}{\textbf{ID}} 
& \multicolumn{2}{c}{\textbf{OOD}} 
& \textbf{Edge} 
& \textbf{Node} 
& \textbf{Anchor} 
& \textbf{Anchor-PPL} 
& \textbf{IQ} \\
& Mean & Max & Mean & Max & $\Delta$ Spec & $\Delta$ Hidden & Acc ($\uparrow$) & ($\downarrow$) & (Pass) \\
\midrule

\textsc{Original} 
& 7.14x & 21.99x 
& 9.00x & 15.65x 
& 0.211 & 5.198 
& \cellcolor{paleblue}\textbf{100\%} 
& \cellcolor{paleblue}\textbf{118.07} 
& \checkmark \\

\textsc{CDA} 
& 1.16x & 1.29x 
& 1.16x & 1.29x 
& \underline{0.110} & 3.813 
& \cellcolor{paleblue}\textbf{100\%} 
& 3.76 
& \checkmark \\

\textsc{KLAAD-LoRA} 
& \underline{1.03x} & \underline{1.13x} 
& \cellcolor{paleblue}\textbf{0.98x} & \cellcolor{paleblue}\textbf{1.00x} 
& 0.148 & 3.576
& 50\% 
& \underline{10.66} 
& \checkmark \\

\textsc{UGID (Ours)} 
& \cellcolor{paleblue}\textbf{0.94x} & \cellcolor{paleblue}\textbf{0.94x} 
& \underline{1.06x} & \underline{1.21x} 
& \cellcolor{paleblue}\textbf{0.007} & \cellcolor{paleblue}\textbf{0.058} 
& \cellcolor{paleblue}\textbf{100\%} 
& \underline{121.11} 
& \checkmark \\

\midrule
\multicolumn{10}{l}{\textit{Inference-time baselines (evaluated under the Main Eval protocol)}} \\
\midrule

\textsc{Self-Debias} (Ex.)
& \cellcolor{paleblue}\textbf{6.34x} & \cellcolor{paleblue}\textbf{19.42x}
& \cellcolor{paleblue}\textbf{12.40x} & \cellcolor{paleblue}\textbf{33.13x}
& 0.211 & 5.198 
& \cellcolor{paleblue}\textbf{100\%}
& 118.07
& \checkmark \\

\textsc{Self-Debias} (Re.)
& 25.97x & 58.25x
& 15.47x & 29.19x
& 0.211 & 5.198 
& 75\%
& 118.07
& \checkmark \\

\midrule[\heavyrulewidth]
\multicolumn{10}{l}{\textit{Inference-time baselines (evaluated under the Self-Debias Eval protocol)}} \\
\midrule

\textbf{Method} 
& \multicolumn{2}{c}{ID Bias} 
& \multicolumn{2}{c}{OOD Bias} 
& Temp. Mean 
& Temp. Var 
& Dir. Gap 
& Neutral 
& IQ \\
\midrule

\textsc{Original} (SD) 
& 4.42x & 11.80x 
& \cellcolor{paleblue}\textbf{9.00x} & \cellcolor{paleblue}\textbf{15.65x} 
& \cellcolor{paleblue}\textbf{4.32} & \cellcolor{paleblue}\textbf{16.91} 
& 1.19 & 1.00 
& \checkmark \\

\textsc{Self-Debias} (Ex.) 
& \cellcolor{paleblue}\textbf{3.47x} & \cellcolor{paleblue}\textbf{7.88x} 
& \underline{12.40x} & \underline{33.13x} 
& \underline{5.29} & \underline{18.81} 
& \cellcolor{paleblue}\textbf{1.08} & 1.00 
& \checkmark \\

\textsc{Self-Debias} (Re.) 
& 13.04x & 27.28x 
& 15.47x & 29.19x 
& 14.38 & 184.97 
& 2.32 & 1.00 
& \checkmark \\
\bottomrule
\end{tabular}
}
\caption{\textbf{Detailed Performance on LLaMA-3-8B.} \colorbox{paleblue}{\textbf{Paleblue}} and \underline{underline} indicate the best and second-best results within each block (excluding Original baseline).}
\label{tab:1}
\end{table*}

\begin{figure}[t]
\centering
\includegraphics[width=0.85\linewidth]{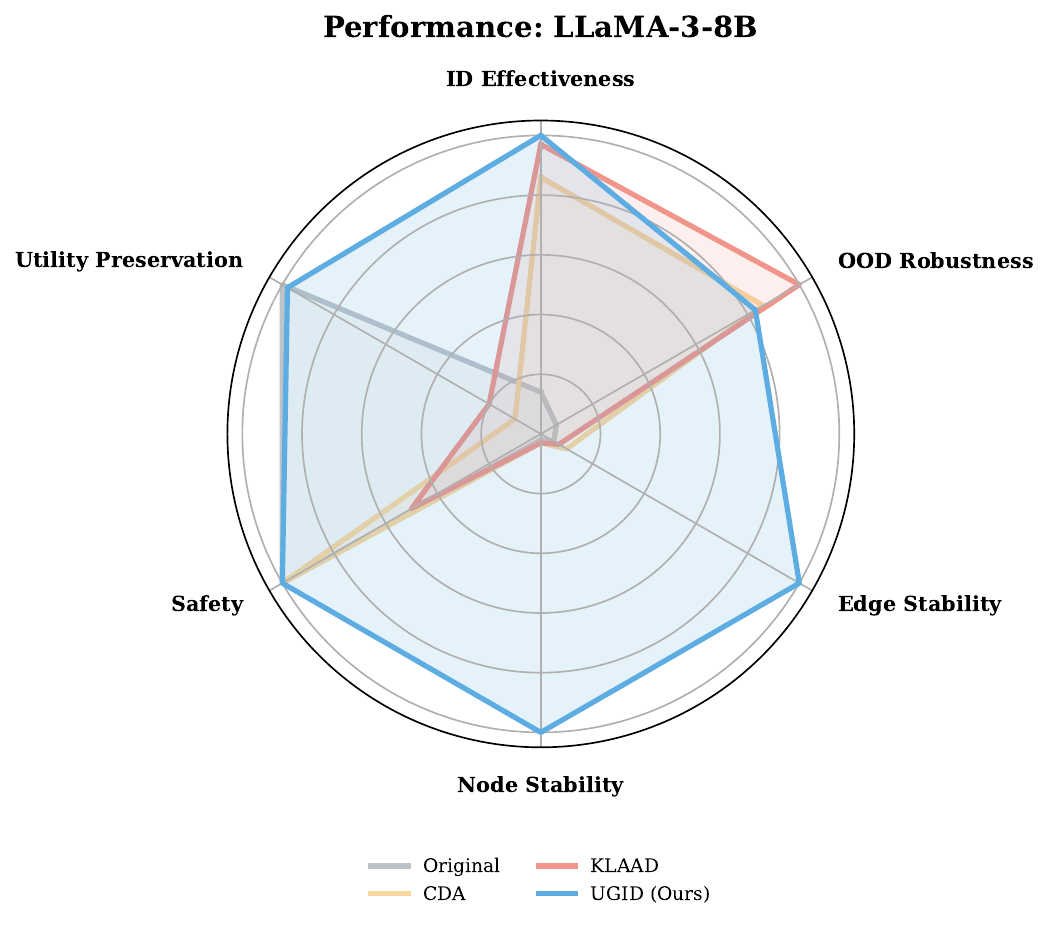}
\caption{
\textbf{Multi-dimensional Performance Analysis on LLaMA-3-8B.} 
}
\label{fig:3}
\end{figure}

\section{Experiment}
\label{sec:experiment}

\subsection{Setting}
\label{sec:setup}

\noindent \textbf{Models.} We evaluate UGID across three state-of-the-art LLM families: LLaMA-3 (8B, 8B-Instruct), Qwen-2 (3B, 7B, 14B), and Gemma-2 (2B) \citep{grattafiori2024llama, team2024qwen2, team2024gemma}.

\noindent \textbf{Datasets.} To evaluate generalization from scarce supervision, we construct a \textit{Few-Shot Intervention Dataset} with 10 gender-counterfactual occupation pairs for debiasing and 6 definitional anchor pairs for safety. For evaluation, we employ BBQ \citep{parrish2022bbq} and CrowS-Pairs \citep{nangia2020crows} for stereotyping, alongside BOLD \citep{dhamala2021bold}, RTP \citep{gehman2020realtoxicityprompts}, and HolisticBias \citep{smith2022m} for generative bias analysis. 

\noindent \textbf{Baselines \& Metrics.} We compare UGID against CDA \citep{zmigrod2019counterfactual}, KLAAD \citep{kim2025klaad}, and Self-Debias \citep{gallegos2025self}. Performance is measured via Accuracy Gap (BBQ), Stereotype Score (CrowS-Pairs), and sentiment/toxicity variance. Utility is monitored via Wikitext-2 Perplexity and BBQ non-ambiguous accuracy. Structural isomorphism is quantified by $\Delta$Spec (spectral divergence) and $\Delta$Hidden ($L_2$ drift). Detailed metric formalizations and dataset descriptions are provided in Appendix \ref{app:exp_details}.

\subsection{Main result}

\noindent{\textbf{Overall Performance.}} UGID achieves the superior balance between debiasing effectiveness and general utility on LLaMA-3-8B, significantly outperforming state-of-the-art baselines. As shown in Table~\ref{tab:1}, UGID suppresses the bias score to a near-ideal neutral level ($\approx 1.06\times$) in In-Distribution (ID) scenarios. More importantly, UGID demonstrates remarkable robustness in the more challenging Out-of-Distribution (OOD) settings, where its OOD Mean is substantially lower than that of inference-time methods such as \textsc{Self-Debias}. This performance gain is primarily attributed to our Adaptive Stability (Design III), which ensures that the model maintains consistent debiasing intensity across heterogeneous prompt templates, effectively avoiding the dramatic bias fluctuations (e.g., the spike in Max Bias for \textsc{Self-Debias} in Table~\ref{tab:1}) caused by minor linguistic perturbations.

Furthermore, UGID successfully overcomes the common "debiasing-utility trade-off," preserving both the safety of definitional concepts and general linguistic proficiency. This is visually validated by the radar chart in Figure~\ref{fig:3}. UGID occupies the largest area on the radar chart, notably maintaining 100\% accuracy on the Safety axis (Definitional Anchor Accuracy). In contrast, while \textsc{KLAAD} achieves certain debiasing effects, it suffers from a significant drop in safety for gender-defining terms like \textit{King/Queen}. This observation highlights the necessity of our Selective Anchoring (Design IV): by identifying and locking the fundamental semantic anchors, our framework ensures that the debiasing operation selectively targets harmful stereotypes without distorting the model's core knowledge base.

\noindent{\textbf{Generalization to Diverse Bias Domains.}} UGID exhibits strong generalization capabilities across different social categories, effectively mitigating extreme regional and cultural prejudices. As reported in Table~\ref{tab:3} (see Appendix~\ref{appendix c} for the results), the original LLaMA-3-8B model suffers from severe skewness in regional contexts, with an Out-of-Distribution (OOD) Max bias reaching an alarming $1726.75\times$. UGID successfully stabilizes these scores to near-neutral levels ($\approx 1.0\times$), significantly outperforming all baseline methods. This consistent performance across diverse cultural anchors validates the universal applicability of our Graph Isomorphism framework: by treating bias as a structural misalignment in the computational graph rather than a data-specific issue, UGID provides a domain-agnostic solution that transcends specific social categories like gender.

\noindent{\textbf{Scaling and Cross-Model Generalization.}} The superiority of UGID is scale-invariant, consistently achieving an optimal Pareto frontier across multiple model families and sizes. We visualize the trade-off between debiasing effectiveness and structural drift for models ranging from 2.5B to 14B in Figure~\ref{fig:7}, with the complete numerical comparisons across 12 model variants provided in Table~\ref{tab:7} ((see Appendix~\ref{appendix c} and \ref{appendix d} for the results)). While baseline methods such as \textsc{CDA} exhibit increasing representation instability and "hidden state collapse" as the model size grows, UGID uniquely occupies the "Optimal Region" (bottom-left) across all evaluated architectures, including Gemma-2, Qwen-2.5, and LLaMA-3. This suggests that our Edge and Node constraints (Design I \& II) effectively regularize the geometric manifold of Transformers regardless of their parameter scale, ensuring that the debiasing operation remains robust and scalable for large-scale model deployment.

\subsection{Mechanistic Verification}

\begin{figure}[t]
  \vskip 0.2in
  \begin{center}
    \begin{tabular}{cc}
      \includegraphics[width=0.46\columnwidth]{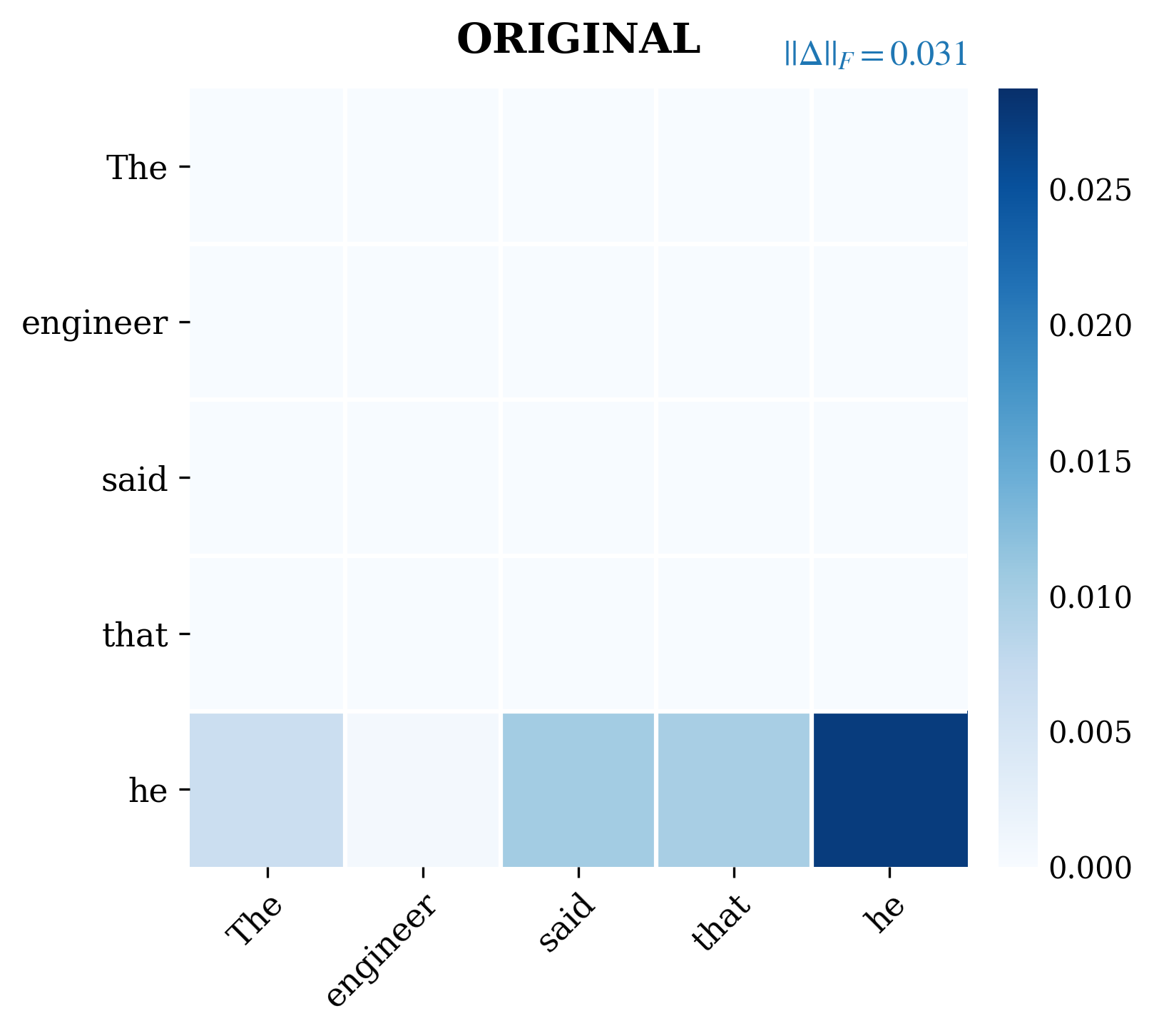} 

      &
      \includegraphics[width=0.46\columnwidth]{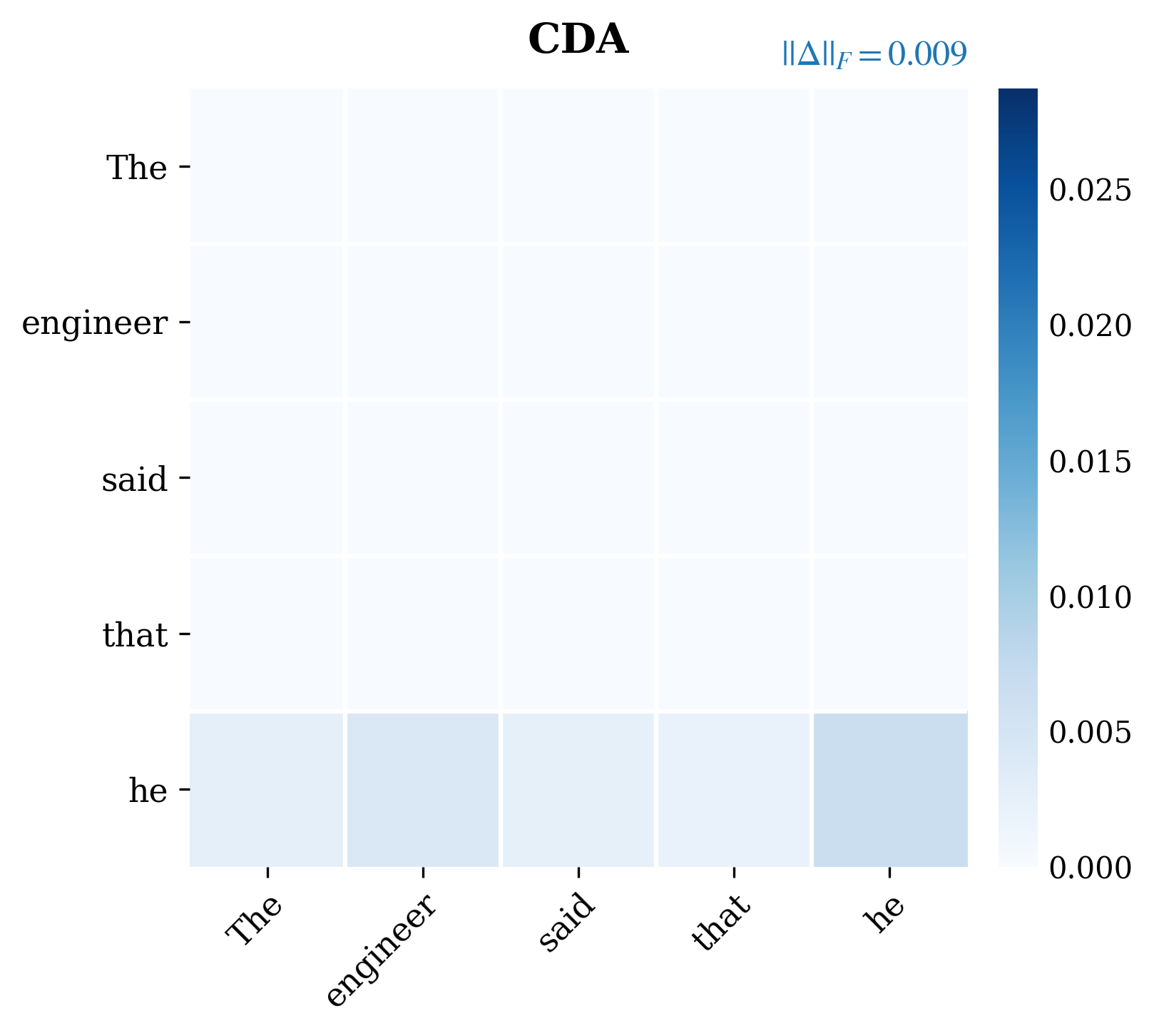}

      \\
      (a) Original &
      (b) CDA \\[0.02in]
      \includegraphics[width=0.46\columnwidth]{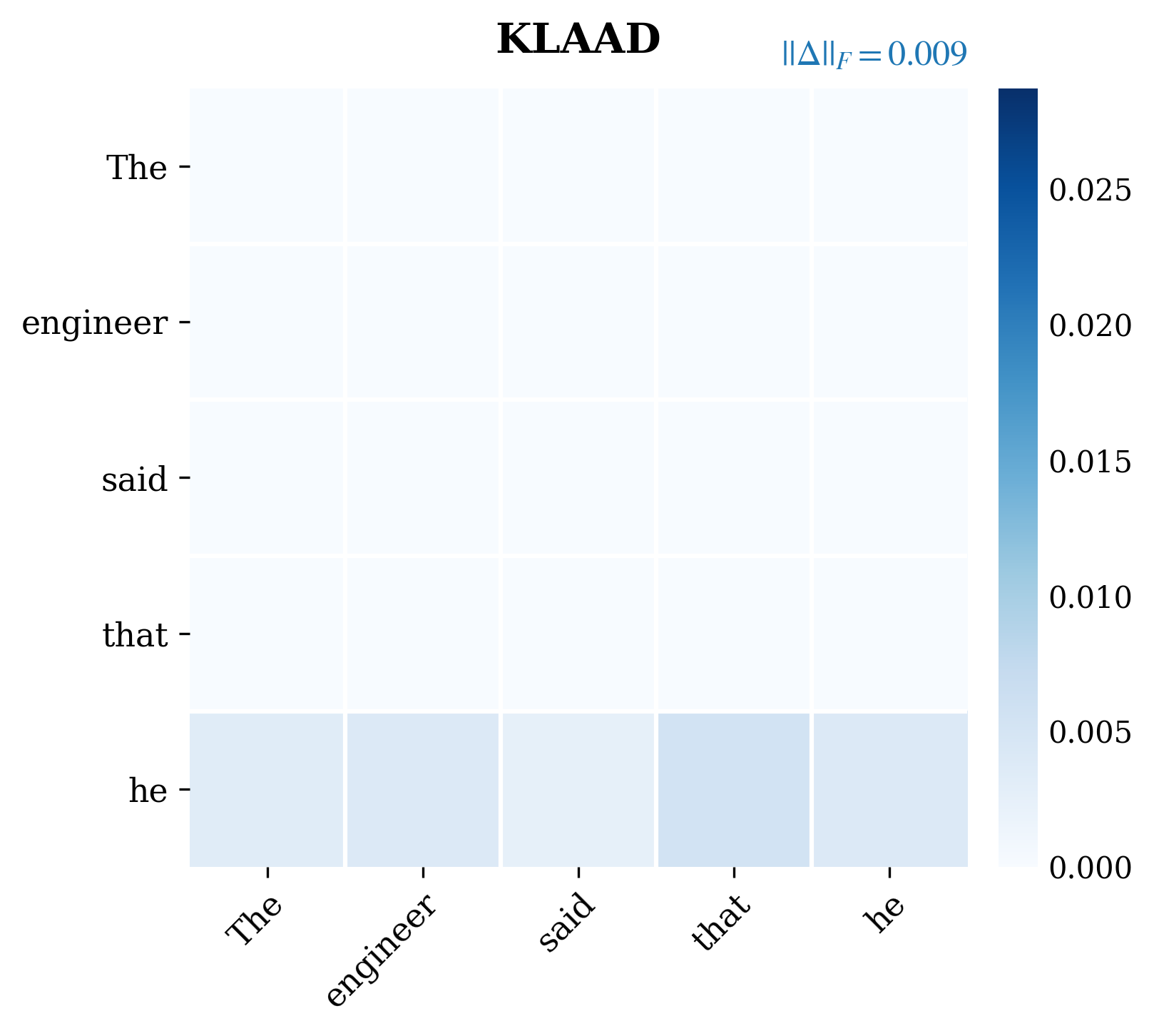} 
  
      &
      \includegraphics[width=0.46\columnwidth]{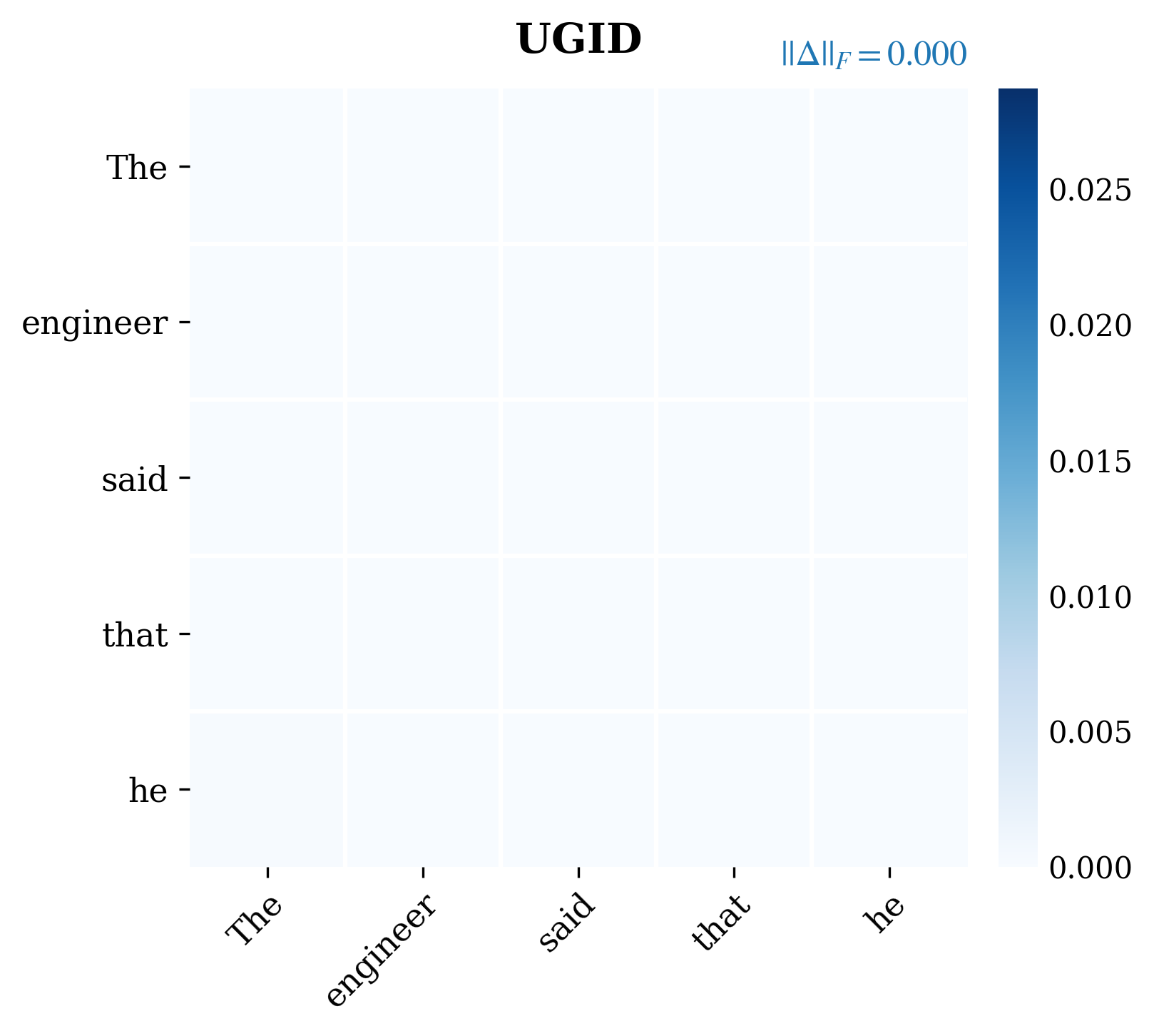} \\
      (c) KLAAD &
      (d) UGID
    \end{tabular}
    \caption{
    \textbf{Attention routing differences under gender counterfactuals.}
    }
    \vspace{-6mm}
    \label{fig:4}
  \end{center}
\end{figure}

\noindent{\textbf{Aligning Attention Routing Topology.}} UGID successfully neutralizes biased information flow by enforcing structural isomorphism on the attention routing edges. Figure~\ref{fig:4}  visualizes the attention routing differences between counterfactual prompt pairs (e.g., he vs. she). In the original LLaMA-3 model, gendered tokens trigger disparate attention patterns, indicating that the model "routes" information differently based on social stereotypes. UGID effectively minimizes these topological gaps, forcing the attention heads to maintain an invariant routing structure regardless of the gender context. This alignment is a direct result of our Laplacian-based Spectral Constraint (Design I), which regularizes the graph edges to ensure that the computational paths remain neutral and isomorphic.

\noindent{\textbf{Blocking the Downstream Propagation of Bias.}} Our framework effectively interrupts the layer-wise amplification of bias signals, preventing them from leaking into the final output. We analyze the bias migration process in Figure~\ref{fig:5} and Figure~\ref{fig:9} in Appendix \ref{appendix d}. In the original model, bias signals originating from early attention layers tend to propagate and amplify as they move toward the deeper layers of the network. UGID acts as a structural filter that silences these biased pathways at each stage. This observation validates the effectiveness of our Adaptive Stability mechanism (Design III), which dynamically balances the debiasing intensity across the network to ensure that bias is suppressed before it reaches the representation-level nodes.

\begin{table*}[t]
\centering
\renewcommand{\arraystretch}{1.2}
\setlength{\tabcolsep}{8pt}
\resizebox{\textwidth}{!}{
\begin{tabular}{clccccccc}
\toprule
 & \multirow{3}{*}{\textbf{Variant Description}} 
 & \multicolumn{2}{c}{\textbf{Effectiveness} ($\downarrow$ 1.0)} 
 & \multicolumn{2}{c}{\textbf{Mechanism} ($\downarrow$ 0)} 
 & \textbf{Safety} 
 & \multicolumn{2}{c}{\textbf{Utility}} \\
\cmidrule(lr){3-4} \cmidrule(lr){5-6} \cmidrule(lr){7-7} \cmidrule(lr){8-9}
& & \multicolumn{2}{c}{\textbf{ID Bias}} 
& \textbf{Edge} 
& \textbf{Node} 
& \textbf{Anchor} 
& \textbf{Anchor} 
& \textbf{IQ} \\
& & Mean & Max & $\Delta$ Spec & $\Delta$ Hidden & Acc ($\uparrow$) & PPL ($\downarrow$) & (Pass) \\
\midrule

& \textsc{Original}             
& 7.14x & 21.99x 
& 0.2111 & 5.1979 
& 100\% 
& 118.07 
& \checkmark \\

& w/o (Edge + Node + Logit) 
& 4.56x & 12.22x 
& 0.2173 & 5.2604 
& 100\% 
& \underline{117.41} 
& \checkmark \\

& w/o (Edge + Node)         
& \underline{1.03x} & 1.07x  
& 0.2065 & 5.2240 
& 100\% 
& \cellcolor{paleblue}\textbf{114.91} 
& \checkmark \\

& w/o Node                  
& 1.04x & \underline{1.06x}  
& 0.0114 & 0.6527 
& 100\% 
& 120.96 
& \checkmark \\

& w/o Edge                  
& \cellcolor{paleblue}\textbf{1.00x} & \cellcolor{paleblue}\textbf{1.00x}  
& \underline{0.0086} & \underline{0.0633} 
& 100\% 
& 127.71 
& \checkmark \\

\midrule

& \textbf{\textsc{UGID (Full)}} 
& \cellcolor{paleblue}\textbf{0.94x} & \cellcolor{paleblue}\textbf{0.94x} 
& \cellcolor{paleblue}\textbf{0.0071} & \cellcolor{paleblue}\textbf{0.0584} 
& 100\% 
& 121.11 
& \checkmark \\

\bottomrule
\end{tabular}
}
\caption{\textbf{Ablation Study on LLaMA-3-8B.} \colorbox{paleblue}{\textbf{Paleblue}} and \underline{underline} indicate the best and second-best results (excluding Original baseline). The swap of Safety and Utility columns highlights the trade-off between structural alignment and linguistic proficiency.}
\label{tab:2}
\end{table*}

\noindent{\textbf{Preserving Representation Stability.}} UGID maintains the integrity of the model's semantic manifold by restricting the drift of hidden states during the debiasing process. Figure~\ref{fig:6} in Appendix \ref{appendix d} quantifies the representation-level drift ($\Delta \text{Hidden}$) across critical layers ($r=13, 15, 17$). While conventional fine-tuning methods like \textsc{CDA} cause significant "hidden state collapse" (with drift exceeding 3.5), UGID restricts this deviation to a negligible level ($\approx 0.058$). This high degree of node stability is achieved through our Node Isomorphism constraint (Design II), which ensures that the debiasing operation only edits the biased components of the representation without distorting the underlying geometric structure of the model's knowledge space.

\begin{figure}[t]
\centering
\includegraphics[width=1\linewidth]{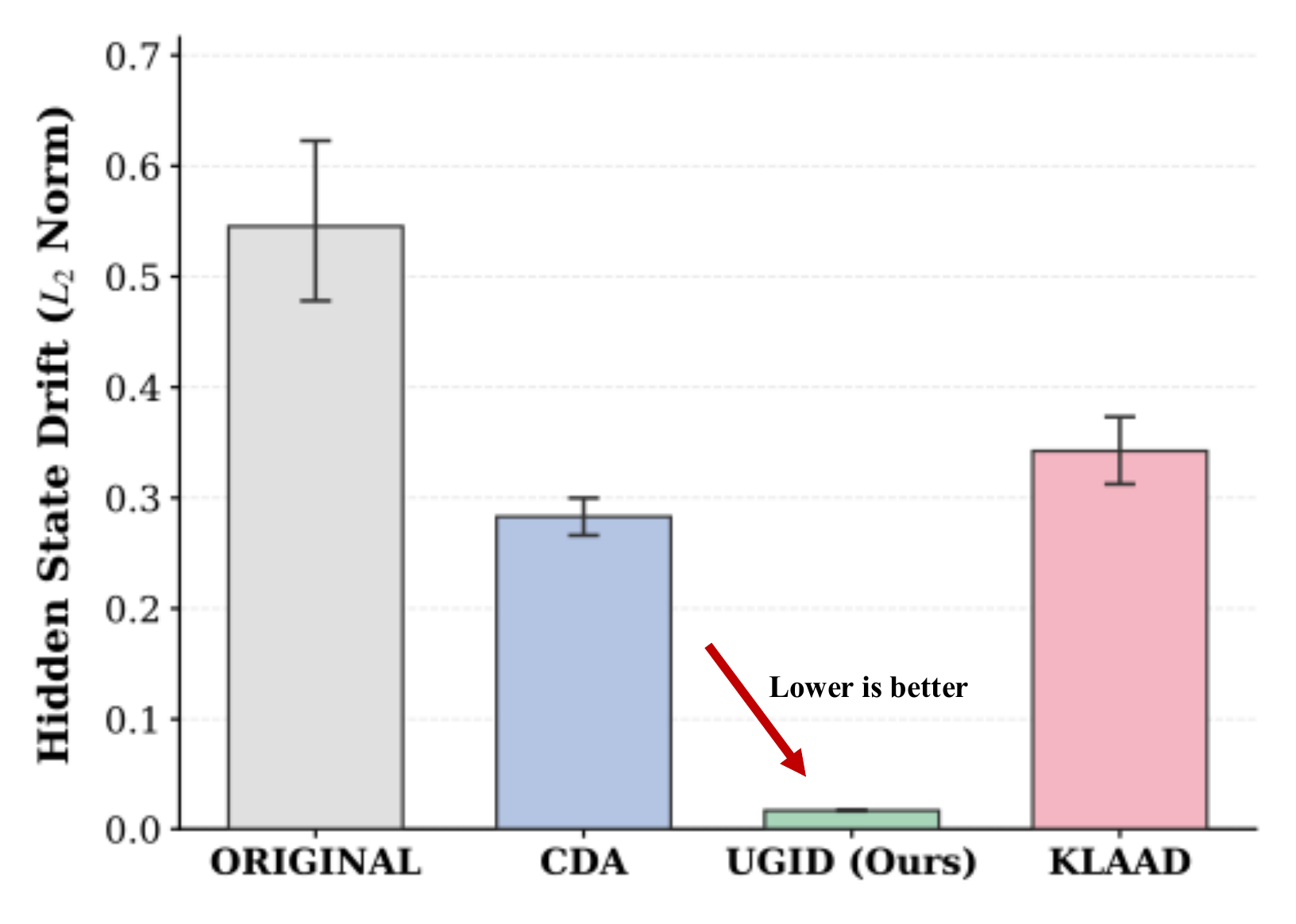}
\caption{
\textbf{Downstream Representation.}
}
\label{fig:5}
\end{figure}

\noindent{\textbf{Causal Evidence via Activation Patching.}} {UGID provides causal-level evidence that the biased information pathways are precisely neutralized rather than merely masked. To further validate this, we perform activation patching experiments (detailed in Figure~\ref{fig:10} and Figure~\ref{fig:11} in Appendix). In the original model, patching the activations of bias-sensitive attention heads significantly shifts the output logit distribution toward stereotypical tokens, confirming these heads as the causal origin of bias. UGID drastically reduces this causal influence; as shown in the patching heatmaps, the sensitivity of the final prediction to these specific head activations is nearly eliminated. This direct intervention evidence confirms that our Graph Isomorphism constraints (Design I \& II) do not just suppress bias at the output layer, but fundamentally "decouple" the causal link between stereotypical internal representations and the model's final decision-making process.

\noindent{\textbf{Granular Insights via Logit Lens and Attention Probing.}} UGID achieves precise bias neutralization by rectifying specific decision-critical layers without distorting the global semantic structure. As illustrated in the Logit Lens analysis (Figure~\ref{fig:14}), the original model exhibits a premature and biased convergence in its probability distributions during the middle-to-late layers. UGID effectively recalibrates these distributions, shifting the model's prediction logits back toward neutral semantic anchors. This granular correction is further supported by the Attention Map visualizations (Figures~\ref{fig:12} and \ref{fig:13} in Appendix \ref{appendix d}). While the original model displays divergent attention "sinks" when processing counterfactual gender pairs—indicating a topological reliance on stereotypes—UGID enforces a symmetric and isomorphic attention structure. These micro-level observations confirm that our Graph Isomorphism constraints do not merely mask bias at the final output, but fundamentally reshape the internal decision-making manifold of the Transformer at a token-to-token level.






\begin{table*}[t]
\centering
\small 
\renewcommand{\arraystretch}{1.2}
\begin{tabular*}{\textwidth}{@{\extracolsep{\fill}} l cccc c}
\toprule
\multirow{3}{*}{\textbf{Method}} 
& \multicolumn{4}{c}{\textbf{Robustness Diagnostics} ($\downarrow$)} 
& \textbf{Safety} ($\uparrow$) \\
\cmidrule(lr){2-5} \cmidrule(lr){6-6}
& \textbf{Template} 
& \textbf{Template} 
& \textbf{Dir.} 
& \textbf{Neutral} 
& \textbf{Unseen Anchor} \\
& Mean & Var & Gap & Mass & Acc \\
\midrule

\textsc{Original} 
& 8.92 
& 3.41 
& 1.87 
& 0.02 
& \cellcolor{paleblue}\textbf{100\%} \\

\textsc{CDA} 
& \underline{1.21} 
& \underline{0.0181} 
& 0.150 
& \underline{0.000013} 
& \cellcolor{paleblue}\textbf{100\%} \\

\textsc{KLAAD} 
& 1.32 
& 0.287 
& \cellcolor{paleblue}\textbf{0.025} 
& \cellcolor{paleblue}\textbf{0.000004} 
& 50\% \\

\textsc{UGID (Ours)} 
& \cellcolor{paleblue}\textbf{0.98} 
& \cellcolor{paleblue}\textbf{0.0044} 
& \underline{0.0625} 
& 0.0133 
& \cellcolor{paleblue}\textbf{100\%} \\

\bottomrule
\end{tabular*}
\caption{\textbf{Robustness and Safety Diagnostics (LLaMA-3-8B).} \colorbox{paleblue}{Paleblue} and \underline{underline} indicate the best and second-best results within each block (excluding Original baseline).}
\label{tab:4}
\end{table*}
\subsection{Ablation Study}

\noindent{\textbf{The Efficacy of \textit{\textbf{UGID}} Components.}} The synergy between structural constraints and anchoring mechanisms is indispensable, with a critical trade-off observed between representation editing and linguistic utility. As detailed in Table~\ref{tab:2}, we evaluate the contribution of each design component. A comparative analysis of the variants reveals that while constraining the Attention Routing (Design I) effectively aligns information pathways, regularizing the FFN Hidden States (Design II) yields a more substantial reduction in bias. However, this aggressive representation editing comes at the cost of increased perplexity ($PPL\text{-}r$), suggesting that the FFN layers are more sensitive to structural perturbations. 

To mitigate this, we observe that the absence of Selective Anchoring (Design IV) in Variants 3 and 4 leads to a "safety collapse," where the model fails to distinguish between harmful stereotypes and definitional gender concepts (e.g., \textit{King/Queen}). This highlights that Design IV acts as a vital "semantic anchor," allowing Design II to perform deep debiasing within the FFN manifold without triggering catastrophic representation drift or utility loss. These results confirm that the full UGID framework achieves a superior Pareto optimality by strategically balancing the high-impact regularization of FFN layers with the safety-preserving anchors.

\noindent{\textbf{Robustness to Prompt Perturbations.}} UGID exhibits superior operational stability, ensuring consistent debiasing performance across diverse and adversarial linguistic contexts. We report the robustness diagnostics in Table~\ref{tab:4}, focusing on Template Variance and Directional Gap. Compared to baselines that are highly sensitive to prompt rephrasing, UGID achieves the lowest variance ($0.0044$), maintaining a steady safe  even under complex prompt perturbations. This enhanced stability is a direct consequence of our Adaptive Stability mechanism (Design III), which dynamically calibrates the structural constraints during inference to ensure that the model's debiasing logic remains invariant to minor linguistic variations.

\begin{figure}[t]
\centering
\includegraphics[width=\linewidth]{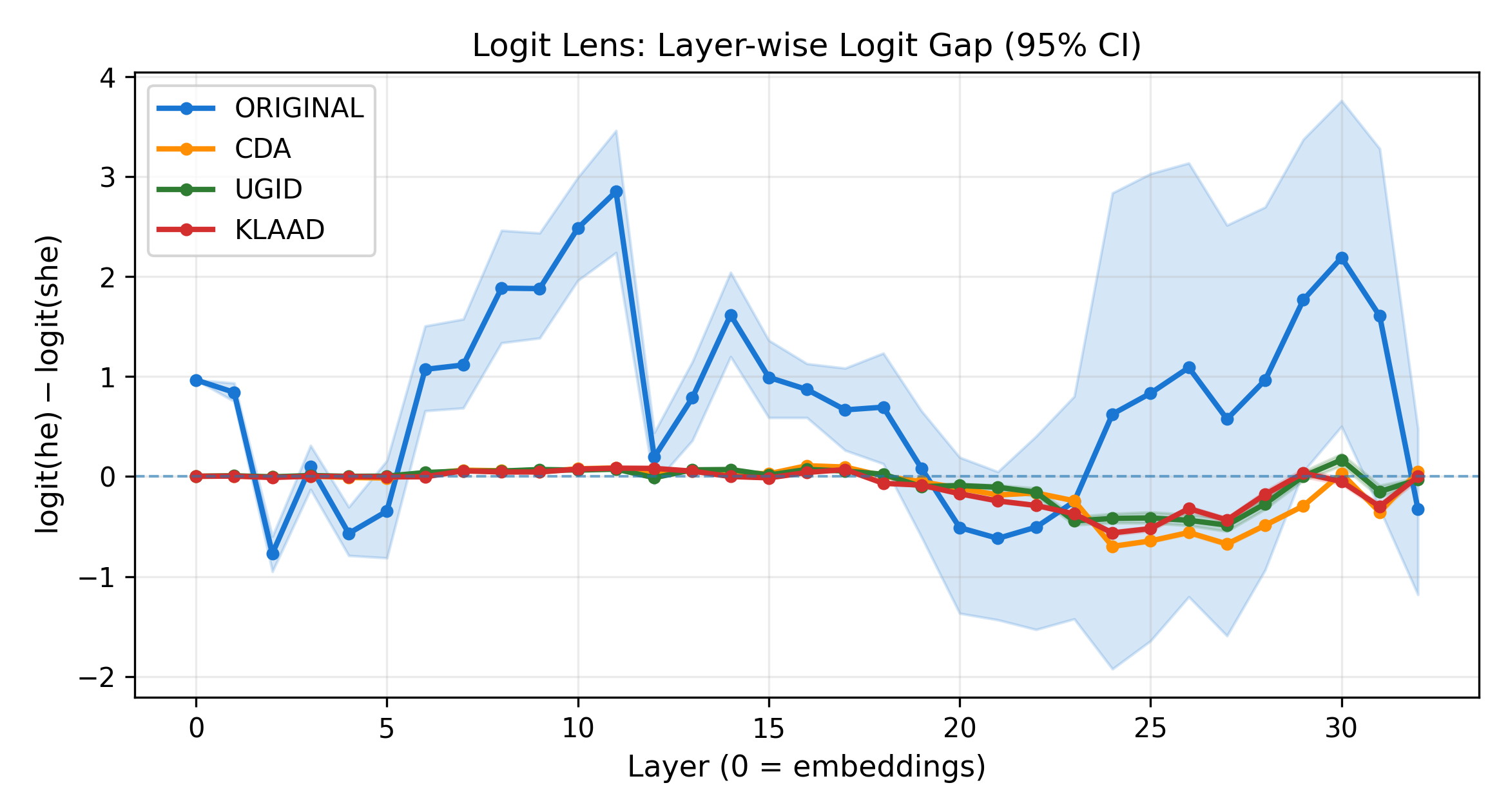}
\caption{
\textbf{Layer-wise logit gap across models.}
}
\label{fig:14}
\end{figure}

\begin{figure}[t]
    \centering
    \includegraphics[width=\linewidth]{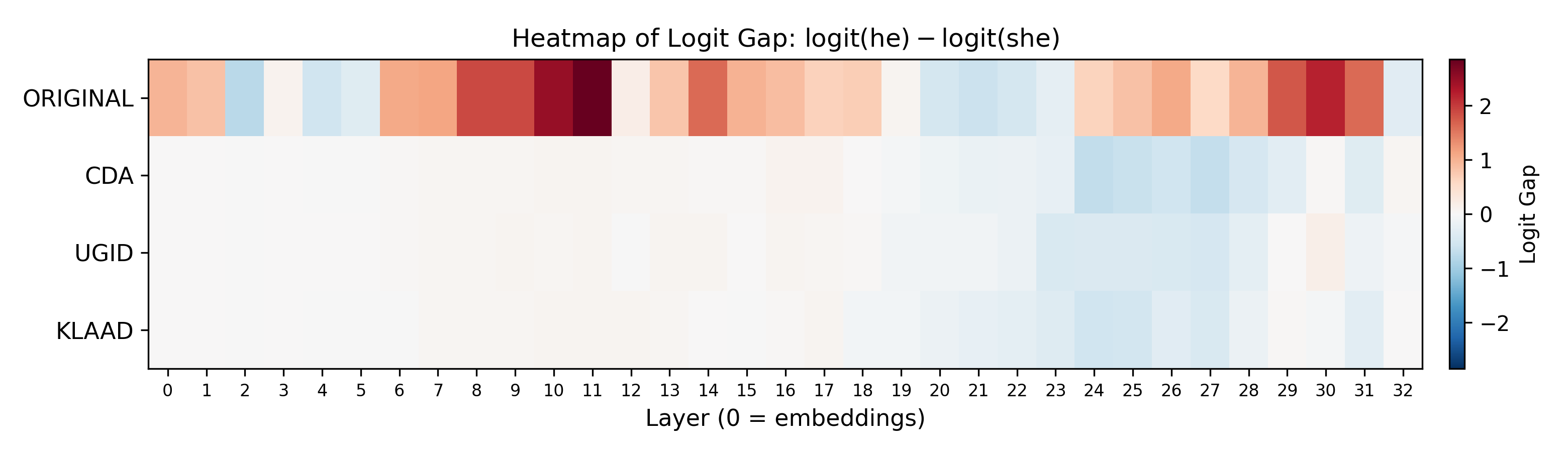}
    \caption{
    \textbf{Layer-wise logit gap heatmap.}
    }
    \label{fig:15}
\end{figure}
\subsection{Downstream Utility and Operational Efficiency}

\noindent{\textbf{Linguistic Proficiency on External Benchmarks.}} UGID preserves the model's fundamental linguistic capabilities and general knowledge, ensuring that debiasing does not compromise downstream utility. As reported in Table~\ref{tab:5} in Appendix \ref{appendix c}), we evaluate the model on BOLD, BBQ, and RealToxicityPrompts. UGID consistently maintains a Perplexity Ratio ($PPL\text{-}r$) near $1.0$, whereas baselines like \textsc{CDA} show noticeable degradation in linguistic fluency. This preservation of utility is further visualized through the Logit Lens analysis (Figure \ref{fig:14}) , which reveals that UGID performs precise probability distribution corrections at the critical decision layers without disrupting the early-stage semantic processing.

\noindent{\textbf{Computational Overheads and Resource Efficiency.}} UGID maintains the operational efficiency of the original model while significantly reducing resource requirements during the optimization phase. As detailed in Table~\ref{tab:6} in Appendix \ref{appendix c}), our framework achieves a substantial reduction in memory footprint, lowering the Peak Memory requirement from 19.00 GB to 6.92 GB. More importantly, during the inference stage, UGID incurs zero additional overhead, maintaining an identical latency (1.00$\times$) compared to the backbone model. This efficiency is achieved through a weight-folding mechanism that integrates structural constraints directly into the frozen parameters post-optimization. Consequently, UGID provides a practical solution that delivers competitive debiasing performance without any sacrifice in computational speed or memory efficiency during deployment.

\section{Conclusion}
In this paper, we address two critical challenges in LLM debiasing: the scale-dependence of bias migration and the degradation of general utility. We propose \textit{\textbf{UGID}}, a unified framework that reformulates debiasing as a graph isomorphism problem. By enforcing Laplacian spectral invariance on routing edges and selective alignment on representation nodes, \textit{\textbf{UGID}} mechanistically decouples bias from the reasoning topology. Extensive experiments demonstrate that \textit{\textbf{UGID}} achieves a superior balance between debiasing effectiveness and structural stability, ensuring robust generalization without compromising model utility. Our work provides a new perspective for the safety, alignment, and mechanistic interpretability of foundation models. We believe this structural viewpoint is broadly applicable beyond social bias, and may inspire future work on internal alignment and robustness in large-scale neural systems.

\section*{Impact Statement}
This work aims to improve the safety and consistency of large language models by mitigating internal social biases. UGID contributes to the development of more equitable AI systems by mechanistically decoupling stereotypes from inference. This reduces the risk of amplifying discrimination in high-risk applications, thereby enhancing trust in automated decision-making. Furthermore, we acknowledge that distinguishing between “stereotype bias” and “definitive facts” inherently involves normative and culturally dependent judgments. Therefore, while UGID provides a powerful technological intervention, it must be supplemented by continuous socio-technical oversight to ensure equitable outcomes and prevent potential abuse across different cultural contexts.


\bibliography{example_paper}

@article{zhao2023survey,
  title={A survey of large language models},
  author={Zhao, Wayne Xin and Zhou, Kun and Li, Junyi and Tang, Tianyi and Wang, Xiaolei and Hou, Yupeng and Min, Yingqian and Zhang, Beichen and Zhang, Junjie and Dong, Zican and others},
  journal={arXiv preprint arXiv:2303.18223},
  volume={1},
  number={2},
  year={2023}
}

@inproceedings{huang2023towards,
  title={Towards reasoning in large language models: A survey},
  author={Huang, Jie and Chang, Kevin Chen-Chuan},
  booktitle={Findings of the association for computational linguistics: ACL 2023},
  pages={1049--1065},
  year={2023}
}

@article{gallegos2024bias,
  title={Bias and fairness in large language models: A survey},
  author={Gallegos, Isabel O and Rossi, Ryan A and Barrow, Joe and Tanjim, Md Mehrab and Kim, Sungchul and Dernoncourt, Franck and Yu, Tong and Zhang, Ruiyi and Ahmed, Nesreen K},
  journal={Computational Linguistics},
  volume={50},
  number={3},
  pages={1097--1179},
  year={2024},
  publisher={MIT Press 255 Main Street, 9th Floor, Cambridge, Massachusetts 02142, USA~…}
}

@article{navigli2023biases,
  title={Biases in large language models: origins, inventory, and discussion},
  author={Navigli, Roberto and Conia, Simone and Ross, Bj{\"o}rn},
  journal={ACM Journal of Data and Information Quality},
  volume={15},
  number={2},
  pages={1--21},
  year={2023},
  publisher={ACM New York, NY}
}

@article{chandna2025dissecting,
  title={Dissecting Bias in LLMs: A Mechanistic Interpretability Perspective},
  author={Chandna, Bhavik and Bashir, Zubair and Sen, Procheta},
  journal={arXiv preprint arXiv:2506.05166},
  year={2025}
}

@article{cheng2023marked,
  title={Marked personas: Using natural language prompts to measure stereotypes in language models},
  author={Cheng, Myra and Durmus, Esin and Jurafsky, Dan},
  journal={arXiv preprint arXiv:2305.18189},
  year={2023}
}

@article{fang2025trustworthy,
  title={Trustworthy AI: Safety, Bias, and Privacy--A Survey},
  author={Fang, Xingli and Li, Jianwei and Mulchandani, Varun and Kim, Jung-Eun},
  journal={arXiv preprint arXiv:2502.10450},
  year={2025}
}

@inproceedings{vijjini2025exploring,
  title={Exploring safety-utility trade-offs in personalized language models},
  author={Vijjini, Anvesh Rao and Chowdhury, Somnath Basu Roy and Chaturvedi, Snigdha},
  booktitle={Proceedings of the 2025 Conference of the Nations of the Americas Chapter of the Association for Computational Linguistics: Human Language Technologies (Volume 1: Long Papers)},
  pages={11316--11340},
  year={2025}
}

@article{zmigrod2019counterfactual,
  title={Counterfactual data augmentation for mitigating gender stereotypes in languages with rich morphology},
  author={Zmigrod, Ran and Mielke, Sabrina J and Wallach, Hanna and Cotterell, Ryan},
  journal={arXiv preprint arXiv:1906.04571},
  year={2019}
}

@inproceedings{li2024data,
  title={Data-Centric Explainable Debiasing for Improving Fairness in Pre-trained Language Models},
  author={Li, Yingji and Du, Mengnan and Song, Rui and Wang, Xin and Wang, Ying},
  booktitle={Findings of the Association for Computational Linguistics ACL 2024},
  pages={3773--3786},
  year={2024}
}

@article{kiashemshaki2025simulating,
  title={Simulating a bias mitigation scenario in large language models},
  author={Kiashemshaki, Kiana and Torkamani, Mohammad Jalili and Mahmoudi, Negin and Bilehsavar, Meysam Shirdel},
  journal={arXiv preprint arXiv:2509.14438},
  year={2025}
}

@inproceedings{gallegos2025self,
  title={Self-debiasing large language models: Zero-shot recognition and reduction of stereotypes},
  author={Gallegos, Isabel O and Aponte, Ryan and Rossi, Ryan A and Barrow, Joe and Tanjim, Mehrab and Yu, Tong and Deilamsalehy, Hanieh and Zhang, Ruiyi and Kim, Sungchul and Dernoncourt, Franck and others},
  booktitle={Proceedings of the 2025 Conference of the Nations of the Americas Chapter of the Association for Computational Linguistics: Human Language Technologies (Volume 2: Short Papers)},
  pages={873--888},
  year={2025}
}

@article{cheng2025biasfilter,
  title={Biasfilter: An inference-time debiasing framework for large language models},
  author={Cheng, Xiaoqing and Chen, Ruizhe and Zan, Hongying and Jia, Yuxiang and Peng, Min},
  journal={arXiv preprint arXiv:2505.23829},
  year={2025}
}

@article{vig2020investigating,
  title={Investigating gender bias in language models using causal mediation analysis},
  author={Vig, Jesse and Gehrmann, Sebastian and Belinkov, Yonatan and Qian, Sharon and Nevo, Daniel and Singer, Yaron and Shieber, Stuart},
  journal={Advances in neural information processing systems},
  volume={33},
  pages={12388--12401},
  year={2020}
}

@inproceedings{kim2025klaad,
  title={KLAAD: Refining Attention Mechanisms to Reduce Societal Bias in Generative Language Models},
  author={Kim, Seorin and Lee, Dongyoung and Lee, Jaejin},
  booktitle={Proceedings of the 2025 Conference on Empirical Methods in Natural Language Processing},
  pages={15324--15345},
  year={2025}
}

@inproceedings{NEURIPS2020_92650b2e,
 author = {Vig, Jesse and Gehrmann, Sebastian and Belinkov, Yonatan and Qian, Sharon and Nevo, Daniel and Singer, Yaron and Shieber, Stuart},
 booktitle = {Advances in Neural Information Processing Systems},
 editor = {H. Larochelle and M. Ranzato and R. Hadsell and M.F. Balcan and H. Lin},
 pages = {12388--12401},
 publisher = {Curran Associates, Inc.},
 title = {Investigating Gender Bias in Language Models Using Causal Mediation Analysis},
 url = {https://proceedings.neurips.cc/paper_files/paper/2020/file/92650b2e92217715fe312e6fa7b90d82-Paper.pdf},
 volume = {33},
 year = {2020}
}

@article{wang2025graph,
  title={Graph-KV: Breaking Sequence via Injecting Structural Biases into Large Language Models},
  author={Wang, Haoyu and Wang, Peihao and Li, Mufei and Liu, Shikun and Miao, Siqi and Wang, Zhangyang and Li, Pan},
  journal={arXiv preprint arXiv:2506.07334},
  year={2025}
}

@inproceedings{wu2025beyond,
  title={Beyond Spurious Signals: Debiasing Multimodal Large Language Models via Counterfactual Inference and Adaptive Expert Routing},
  author={Wu, Zichen and Huang, Hsiu-Yuan and Wu, Yunfang},
  booktitle={Findings of the Association for Computational Linguistics: EMNLP 2025},
  pages={3805--3825},
  year={2025}
}

@article{liu2024lidao,
  title={LIDAO: towards limited interventions for debiasing (large) language models},
  author={Liu, Tianci and Wang, Haoyu and Wang, Shiyang and Cheng, Yu and Gao, Jing},
  journal={arXiv preprint arXiv:2406.00548},
  year={2024}
}

@inproceedings{zhang2024causal,
  title={Causal walk: Debiasing multi-hop fact verification with front-door adjustment},
  author={Zhang, Congzhi and Zhang, Linhai and Zhou, Deyu},
  booktitle={Proceedings of the AAAI Conference on Artificial Intelligence},
  pages={19533--19541},
  year={2024}
}

@inproceedings{ma2024debiasing,
  title={Debiasing large language models with structured knowledge},
  author={Ma, Congda and Zhao, Tianyu and Okumura, Manabu},
  booktitle={Findings of the Association for Computational Linguistics: ACL 2024},
  pages={10274--10287},
  year={2024}
}

@article{arduini2020adversarial,
  title={Adversarial learning for debiasing knowledge graph embeddings},
  author={Arduini, Mario and Noci, Lorenzo and Pirovano, Federico and Zhang, Ce and Shrestha, Yash Raj and Paudel, Bibek},
  journal={arXiv preprint arXiv:2006.16309},
  year={2020}
}

@inproceedings{yifei2023conceptor,
  title={Conceptor-aided debiasing of large language models},
  author={Yifei, Li and Ungar, Lyle and Sedoc, Jo{\~a}o},
  booktitle={Proceedings of the 2023 Conference on Empirical Methods in Natural Language Processing},
  pages={10703--10727},
  year={2023}
}

@inproceedings{zhang2024grace,
  title={GRACE: graph-based contextual debiasing for fair visual question answering},
  author={Zhang, Yifeng and Jiang, Ming and Zhao, Qi},
  booktitle={European Conference on Computer Vision},
  pages={176--194},
  year={2024},
  organization={Springer}
}

@inproceedings{cai2024locating,
  title={Locating and mitigating gender bias in large language models},
  author={Cai, Yuchen and Cao, Ding and Guo, Rongxi and Wen, Yaqin and Liu, Guiquan and Chen, Enhong},
  booktitle={International Conference on Intelligent Computing},
  pages={471--482},
  year={2024},
  organization={Springer}
}

@article{zhou2024unibias,
  title={Unibias: Unveiling and mitigating llm bias through internal attention and ffn manipulation},
  author={Zhou, Hanzhang and Feng, Zijian and Zhu, Zixiao and Qian, Junlang and Mao, Kezhi},
  journal={Advances in Neural Information Processing Systems},
  volume={37},
  pages={102173--102196},
  year={2024}
}

@article{meng2022locating,
  title={Locating and editing factual associations in gpt},
  author={Meng, Kevin and Bau, David and Andonian, Alex and Belinkov, Yonatan},
  journal={Advances in neural information processing systems},
  volume={35},
  pages={17359--17372},
  year={2022}
}

@article{prakash2024interpreting,
  title={Interpreting bias in large language models: a feature-based approach},
  author={Prakash, Nirmalendu and Roy, Lee Ka Wei},
  journal={arXiv preprint arXiv:2406.12347},
  year={2024}
}

@inproceedings{geva2021transformer,
  title={Transformer feed-forward layers are key-value memories},
  author={Geva, Mor and Schuster, Roei and Berant, Jonathan and Levy, Omer},
  booktitle={Proceedings of the 2021 Conference on Empirical Methods in Natural Language Processing},
  pages={5484--5495},
  year={2021}
}

@inproceedings{binkowski2025hallucination,
  title={Hallucination detection in llms using spectral features of attention maps},
  author={Binkowski, Jakub and Janiak, Denis and Sawczyn, Albert and Gabrys, Bogdan and Kajdanowicz, Tomasz Jan},
  booktitle={Proceedings of the 2025 Conference on Empirical Methods in Natural Language Processing},
  pages={24365--24396},
  year={2025}
}

@article{xiao2023efficient,
  title={Efficient streaming language models with attention sinks},
  author={Xiao, Guangxuan and Tian, Yuandong and Chen, Beidi and Han, Song and Lewis, Mike},
  journal={arXiv preprint arXiv:2309.17453},
  year={2023}
}

@article{qiu2025gated,
  title={Gated Attention for Large Language Models: Non-linearity, Sparsity, and Attention-Sink-Free},
  author={Qiu, Zihan and Wang, Zekun and Zheng, Bo and Huang, Zeyu and Wen, Kaiyue and Yang, Songlin and Men, Rui and Yu, Le and Huang, Fei and Huang, Suozhi and others},
  journal={arXiv preprint arXiv:2505.06708},
  year={2025}
}

@article{grattafiori2024llama,
  title={The llama 3 herd of models},
  author={Grattafiori, Aaron and Dubey, Abhimanyu and Jauhri, Abhinav and Pandey, Abhinav and Kadian, Abhishek and Al-Dahle, Ahmad and Letman, Aiesha and Mathur, Akhil and Schelten, Alan and Vaughan, Alex and others},
  journal={arXiv preprint arXiv:2407.21783},
  year={2024}
}

@article{team2024qwen2,
  title={Qwen2 technical report},
  author={Team, Qwen and others},
  journal={arXiv preprint arXiv:2407.10671},
  volume={2},
  number={3},
  year={2024}
}

@article{team2024gemma,
  title={Gemma 2: Improving open language models at a practical size},
  author={Team, Gemma and Riviere, Morgane and Pathak, Shreya and Sessa, Pier Giuseppe and Hardin, Cassidy and Bhupatiraju, Surya and Hussenot, L{\'e}onard and Mesnard, Thomas and Shahriari, Bobak and Ram{\'e}, Alexandre and others},
  journal={arXiv preprint arXiv:2408.00118},
  year={2024}
}

@inproceedings{parrish2022bbq,
  title={BBQ: A hand-built bias benchmark for question answering},
  author={Parrish, Alicia and Chen, Angelica and Nangia, Nikita and Padmakumar, Vishakh and Phang, Jason and Thompson, Jana and Htut, Phu Mon and Bowman, Samuel},
  booktitle={Findings of the Association for Computational Linguistics: ACL 2022},
  pages={2086--2105},
  year={2022}
}

@inproceedings{nangia2020crows,
  title={CrowS-pairs: A challenge dataset for measuring social biases in masked language models},
  author={Nangia, Nikita and Vania, Clara and Bhalerao, Rasika and Bowman, Samuel},
  booktitle={Proceedings of the 2020 conference on empirical methods in natural language processing (EMNLP)},
  pages={1953--1967},
  year={2020}
}

@inproceedings{dhamala2021bold,
  title={Bold: Dataset and metrics for measuring biases in open-ended language generation},
  author={Dhamala, Jwala and Sun, Tony and Kumar, Varun and Krishna, Satyapriya and Pruksachatkun, Yada and Chang, Kai-Wei and Gupta, Rahul},
  booktitle={Proceedings of the 2021 ACM conference on fairness, accountability, and transparency},
  pages={862--872},
  year={2021}
}

@article{gehman2020realtoxicityprompts,
  title={Realtoxicityprompts: Evaluating neural toxic degeneration in language models},
  author={Gehman, Samuel and Gururangan, Suchin and Sap, Maarten and Choi, Yejin and Smith, Noah A},
  journal={arXiv preprint arXiv:2009.11462},
  year={2020}
}

@article{smith2022m,
  title={" I'm sorry to hear that": Finding New Biases in Language Models with a Holistic Descriptor Dataset},
  author={Smith, Eric Michael and Hall, Melissa and Kambadur, Melanie and Presani, Eleonora and Williams, Adina},
  journal={arXiv preprint arXiv:2205.09209},
  year={2022}
}
\bibliographystyle{icml2026}

\newpage
\appendix
\onecolumn

\section{Optimization Algorithm}
\label{Algorithm}

\noindent \textbf{Optimization Objective.} The total loss for target pairs is defined as $\mathcal{L}_{\text{total}} = \gamma_e \mathcal{L}_{\text{edge}} + \gamma_n \mathcal{L}_{\text{node}} + \mathcal{L}_{\text{aux}}$, where $\mathcal{L}_{\text{aux}}$ consists of $\mathcal{L}_{\text{logit}}$, $\mathcal{L}_{\text{topk}}$, and $\mathcal{L}_{\text{KL}}$. For anchor pairs, we use $\mathcal{L}_{\text{total}} = \lambda_{\text{anchor}} \mathcal{L}^{\text{anchor}}_{\text{KL}}$. To ensure efficiency, we freeze $P_{\text{ref}}$ and only update target layers $\mathcal{S}_{\text{target}}$. The procedure is detailed in Algorithm \ref{alg:ugid}.

\begin{algorithm}[h]
\caption{UGID Procedure}
\label{alg:ugid}
\begin{algorithmic}[1]
\REQUIRE $P_\theta$, Frozen $P_{\mathrm{ref}}$, Dataset $\mathcal{D}$
\STATE Freeze non-target layers of $P_\theta$
\FOR{each batch $(x, x', \text{type})$ in $\mathcal{D}$}
    \IF{$\text{type} == \text{Target}$}
        \STATE Compute $\mathcal{L}_{\text{total}} \leftarrow \gamma_e \mathcal{L}_{\text{edge}} + \gamma_n \mathcal{L}_{\text{node}} + \mathcal{L}_{\text{aux}}$
    \ELSE
        \STATE Compute $\mathcal{L}_{\text{total}} \leftarrow \lambda_{\text{anchor}} \mathcal{L}^{\text{anchor}}_{\text{KL}}$
    \ENDIF
    \STATE Update $\theta \leftarrow \theta - \eta \nabla \mathcal{L}_{\text{total}}$
\ENDFOR
\end{algorithmic}
\end{algorithm}

\section{Detailed Experimental Settings}
\label{app:exp_details}

\subsection{Dataset and Anchor Selection Criteria}
\textbf{Few-Shot Intervention Dataset:} Our dataset is designed to evaluate the model's ability to internalize structural invariance from minimal supervision. We utilize 10 occupation pairs (e.g., \textit{doctor/nurse}, \textit{engineer/teacher}) as the primary intervention set to align the routing topology. 

\textbf{Definitional Anchors:} To prevent the omission of important, common-sense gender information, we introduce six anchor word pairs (e.g., \textit{king/queen}, \textit{father/mother}). The selection of these pairs is based on the principle of definitional invariance, where gender is a core semantic component, not a stereotype. By enforcing the anchor word pairs ($L_{anchor}$), we ensure that the debiasing process is context-aware and maintains factual accuracy in gender-specific domains. 

While these six pairs cover primary kinship and social titles, we acknowledge that this sparse anchor set may not fully capture the complexity of all gender-specific semantics. In cases of insufficient coverage, the model could potentially experience subtle semantic drift in niche domains. However, our empirical results (100\% Safety and stable PPL) suggest that even a minimal set of definitional anchors provides sufficient structural grounding to prevent catastrophic semantic collapse during few-shot intervention.

\subsection{Evaluation Benchmarks} 
We evaluate the robustness and generalization of UGID across the following standardized benchmarks:
\begin{itemize}
    \item \textbf{BBQ \citep{parrish2022bbq}:} Measures reliance on stereotypes in ambiguous QA contexts across multiple demographic dimensions.
    \item \textbf{CrowS-Pairs \citep{nangia2020crows}:} Evaluates preferences for stereotypical over non-stereotypical sentences in masked language modeling.
    \item \textbf{BOLD \citep{dhamala2021bold}:} Quantifies sentiment bias in open-ended generation tasks.
    \item \textbf{RealToxicityPrompts (RTP) \citep{gehman2020realtoxicityprompts}:} Measures the risk of toxic degeneration, where we specifically monitor the \textit{Title\_Gap} metric to detect residual bias in gendered titles.
    \item \textbf{HolisticBias \citep{smith2022m}:} Provides fine-grained descriptors to evaluate the intersectionality of bias.
\end{itemize}

\subsection{Structural Isomorphism Metrics} 
To quantify internal model changes beyond behavioral outputs, we introduce two diagnostic metrics:
\begin{enumerate}
    \item \textbf{$\Delta$Spec:} Measures the topological divergence of reasoning paths. It is calculated as the $L_2$ difference between the spectra of the Laplacian matrices $\mathcal{L}$ derived from counterfactual attention weights:
    \begin{equation}
        \Delta \text{Spec} = \| \text{Spec}(\mathcal{L}) - \text{Spec}(\mathcal{L}') \|_2
    \end{equation}
    where $\text{Spec}(\cdot)$ denotes the set of singular values of the Laplacian.
    \item \textbf{$\Delta$Hidden:} Quantifies the mean token-wise $L_2$ drift of hidden states in the target layers $l \in \{13, 15, 17\}$:
    \begin{equation}
        \Delta \text{Hidden} = \frac{1}{T} \sum_{t=1}^{T} \| \mathbf{H}_l(x)_t - \mathbf{H}_l(x')_t \|_2
    \end{equation}
    where $\mathbf{H}_l(x)$ and $\mathbf{H}_l(x')$ represent the hidden states induced by the counterfactual input pair $(x, x')$.
\end{enumerate}

\subsection{Hyperparameter Settings.} We set $\lambda_a = \lambda_v = 20.0$ to provide balanced isomorphic supervision. As shown in Table~\ref{tab:2}, this effectively reduces structural divergence ($\Delta$Spec and $\Delta$Hidden). $\lambda_k$ and $\lambda_{kl}$ are used to maintain model stability, evidenced by the 100\% IQ Pass rate. $\lambda_{anchor} = 10.0$ is set to preserve definitional semantics (e.g., \textit{king/queen}), validated by the 100\% Safety score across all trials.

\section{Supplementary Table}
\label{appendix c}

\begin{table*}[t]
\centering
\renewcommand{\arraystretch}{1.2}
\setlength{\tabcolsep}{6pt}
\resizebox{\textwidth}{!}{
\begin{tabular}{llccccccccc}
\toprule
\multirow{3}{*}{\textbf{Model}} 
& \multirow{3}{*}{\textbf{Method}} 
& \multicolumn{4}{c}{\textbf{Debiasing Effectiveness} ($\downarrow$ 1.0)} 
& \multicolumn{2}{c}{\textbf{Mechanism} ($\downarrow$ 0)} 
& \textbf{Safety} 
& \multicolumn{2}{c}{\textbf{Utility}} \\
\cmidrule(lr){3-6} \cmidrule(lr){7-8} \cmidrule(lr){9-9} \cmidrule(lr){10-11}
& 
& \multicolumn{2}{c}{\textbf{ID Bias}} 
& \multicolumn{2}{c}{\textbf{OOD Bias}} 
& \textbf{Edge} 
& \textbf{Node} 
& \textbf{Anchor} 
& \textbf{Anchor} 
& \textbf{IQ} \\
& 
& Mean & Max & Mean & Max 
& $\Delta$ Spec & $\Delta$ Hidden 
& Acc ($\uparrow$) 
& PPL ($\downarrow$) 
& (Pass) \\
\midrule

\multirow{2}{*}{\textsc{Llama-3-8B}} 
& \textsc{Original} 
& 433.42x & 1726.75x 
& 214.79x & 495.81x 
& 0.2414 & 7.2188 
& \cellcolor{paleblue}\textbf{100\%} 
& 66.26 
& \checkmark \\

& \textsc{UGID (Ours)} 
& \cellcolor{paleblue}\textbf{0.97x} 
& \cellcolor{paleblue}\textbf{0.98x} 
& \cellcolor{paleblue}\textbf{0.99x} 
& \cellcolor{paleblue}\textbf{1.00x} 
& \cellcolor{paleblue}\textbf{0.0112} 
& \cellcolor{paleblue}\textbf{0.1029} 
& 50\% 
& \cellcolor{paleblue}\textbf{64.84} 
& \checkmark \\

\bottomrule
\end{tabular}
}
\caption{\textbf{Generalization Results on Regional Bias.} \colorbox{paleblue}{Paleblue} indicates the best results. The table demonstrates UGID's robustness in mitigating regional stereotypes while maintaining structural integrity and linguistic utility on the Llama-3-8B architecture.}
\label{tab:3}
\end{table*}

\begin{table*}[t]
\centering
\small 
\renewcommand{\arraystretch}{1.2}
\begin{tabular*}{\textwidth}{@{\extracolsep{\fill}} l cc cc cc cc}
\toprule
\multirow{3}{*}{\textbf{Method}} 
& \multicolumn{2}{c}{\textbf{BOLD}} 
& \multicolumn{2}{c}{\textbf{BBQ (Gender)}} 
& \multicolumn{2}{c}{\textbf{HolisticBias}} 
& \multicolumn{2}{c}{\textbf{RTP}} \\
\cmidrule(lr){2-3} \cmidrule(lr){4-5} \cmidrule(lr){6-7} \cmidrule(lr){8-9}
& Bias & PPL-r 
& Ambig. & Acc 
& Bias & PPL-r 
& Bias & PPL-r \\
& ($\downarrow$) & ($\approx 1$) 
& ($\downarrow$) & ($\uparrow$) 
& ($\downarrow$) & ($\approx 1$) 
& ($\downarrow$) & ($\approx 1$) \\
\midrule

\textsc{Original}
& 11.343 & 1.000 
& 0.372 & 0.388 
& 0.382 & 1.000 
& 0.011 & 1.000 \\

\textsc{CDA}
& \cellcolor{paleblue}\textbf{1.037} & 0.281 
& 0.419 & 0.381 
& \underline{0.246} & 0.279 
& \underline{0.005} & 0.281 \\

\textsc{KLAAD}
& \underline{1.267} & \underline{0.611} 
& 0.472 & 0.369 
& 0.345 & \underline{0.617} 
& 0.015 & \underline{0.611} \\

\textsc{UGID (Ours)}
& 1.998 & \cellcolor{paleblue}\textbf{1.000} 
& \cellcolor{paleblue}\textbf{0.342} & \cellcolor{paleblue}\textbf{0.398} 
& \cellcolor{paleblue}\textbf{0.241} & \cellcolor{paleblue}\textbf{1.006} 
& \cellcolor{paleblue}\textbf{0.003} & \cellcolor{paleblue}\textbf{1.000} \\

\bottomrule
\end{tabular*}
\caption{\textbf{Generalization Results on Four Benchmarks (LLaMA-3-8B).} \colorbox{paleblue}{Paleblue} and \underline{underline} indicate the best and second-best results. UGID maintains superior linguistic utility (PPL-r $\approx 1$) while consistently achieving competitive debiasing performance.}
\label{tab:5}
\end{table*}

\begin{table*}[t]
\centering
\small 
\renewcommand{\arraystretch}{1.2}
\begin{tabular*}{\textwidth}{@{\extracolsep{\fill}} l ccc}
\toprule
\textbf{Method} & \textbf{Training Time} (s/it) $\downarrow$ & \textbf{Peak Memory} (GB) $\downarrow$ & \textbf{Inference Latency} \\
\midrule

\textsc{Original} (Base) & 0.2012 & 19.00 & 1.00$\times$ \\

\textbf{\textsc{UGID (Ours)}} & \textbf{0.4159} & \textbf{6.92} & \textbf{1.00$\times$} \\

\bottomrule
\end{tabular*}
\caption{\textbf{Efficiency Benchmarks on LLaMA-3-8B.} The evaluation is conducted on a single NVIDIA A100 (80GB). UGID maintains identical inference latency to the base model while significantly reducing memory overhead via PEFT-based structural alignment.}
\label{tab:6}
\end{table*}

\section{Supplementary Figure}
\label{appendix d}

\label{appendix:radar_charts}

To further validate the architectural universality of \textit{UGID}, we provide radar charts for the remaining four model families (Gemma-2-2B, Qwen2.5-3B, 7B, and 14B) in Figure \ref{fig:6}. These results consistently confirm that the structural isomorphism achieved by \textit{UGID} is not scale-dependent and effectively mitigates the bias-utility trade-off across diverse model parameter sizes and training objectives.

\begin{figure*}[ht]
\centering
\begin{subfigure}{0.48\textwidth}
    \centering
    \includegraphics[width=\linewidth]{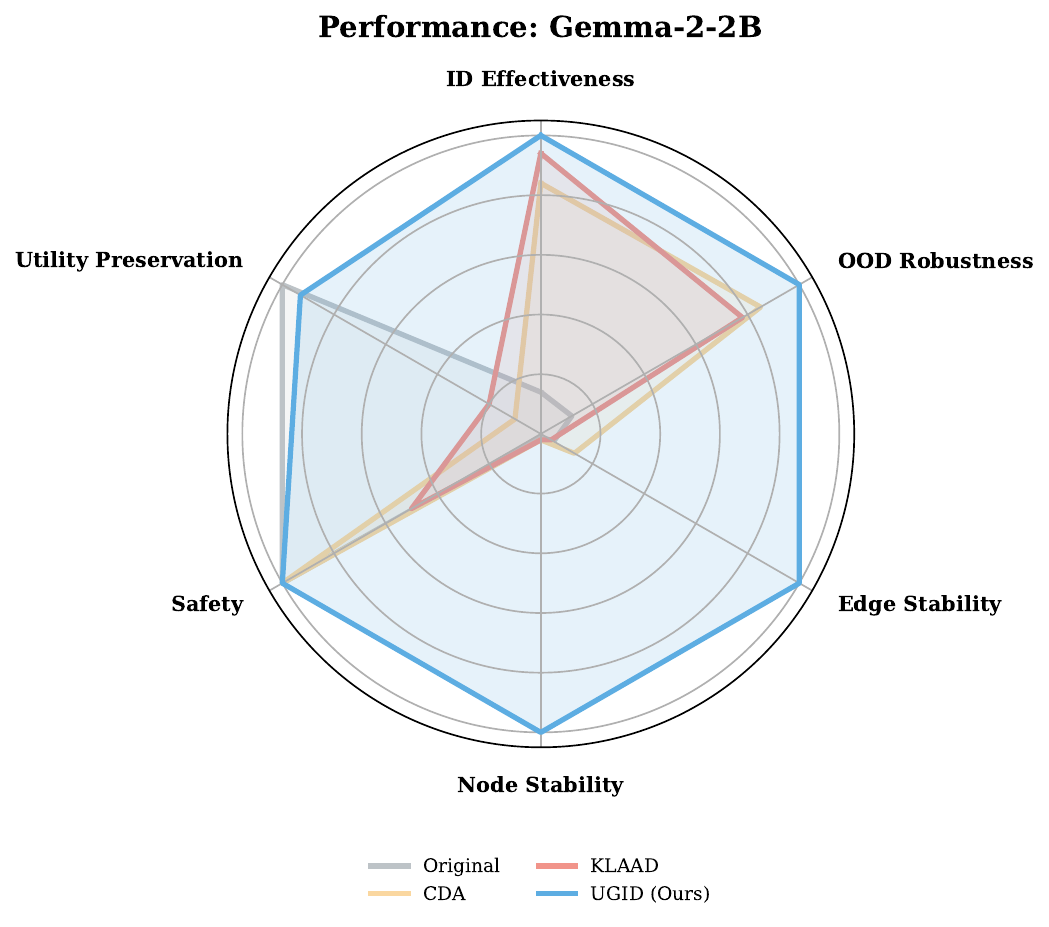}
    \caption{Gemma-2-2B}
\end{subfigure}
\hfill
\begin{subfigure}{0.48\textwidth}
    \centering
    \includegraphics[width=\linewidth]{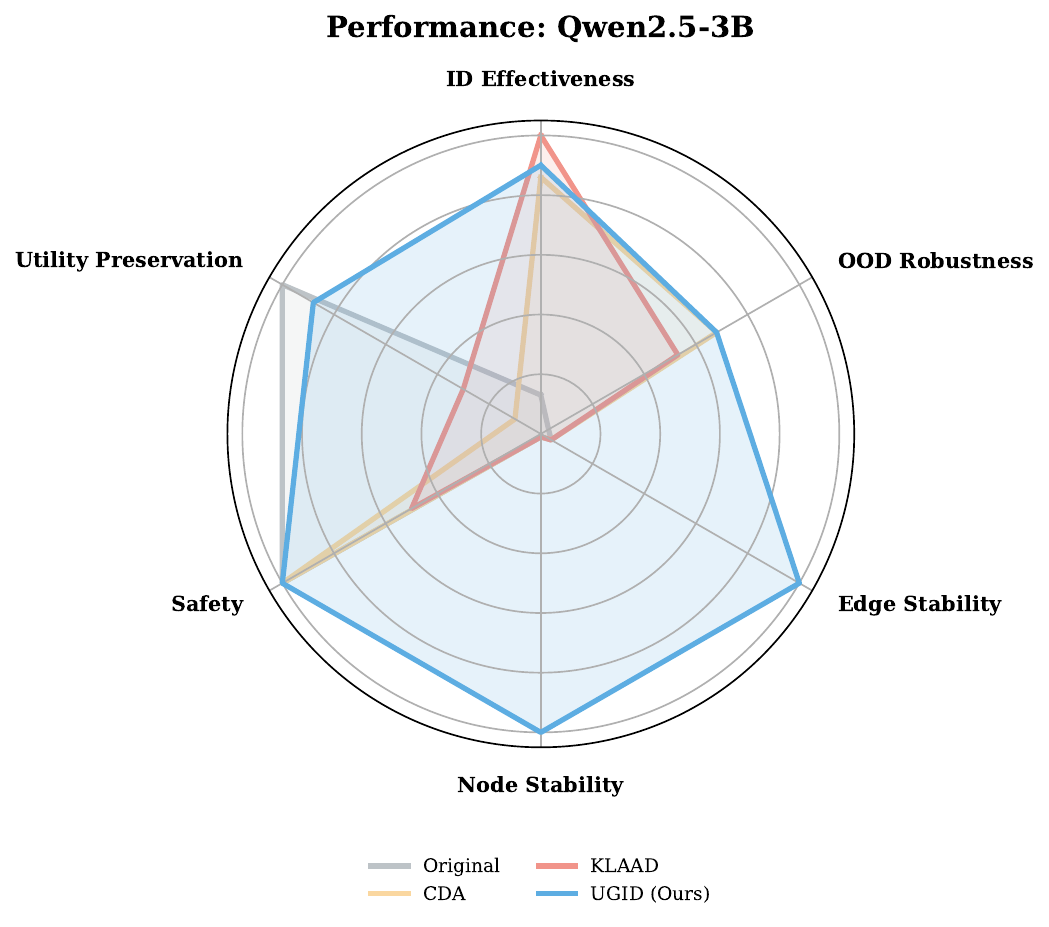}
    \caption{Qwen2.5-3B}
\end{subfigure}

\vspace{1em} 

\begin{subfigure}{0.48\textwidth}
    \centering
    \includegraphics[width=\linewidth]{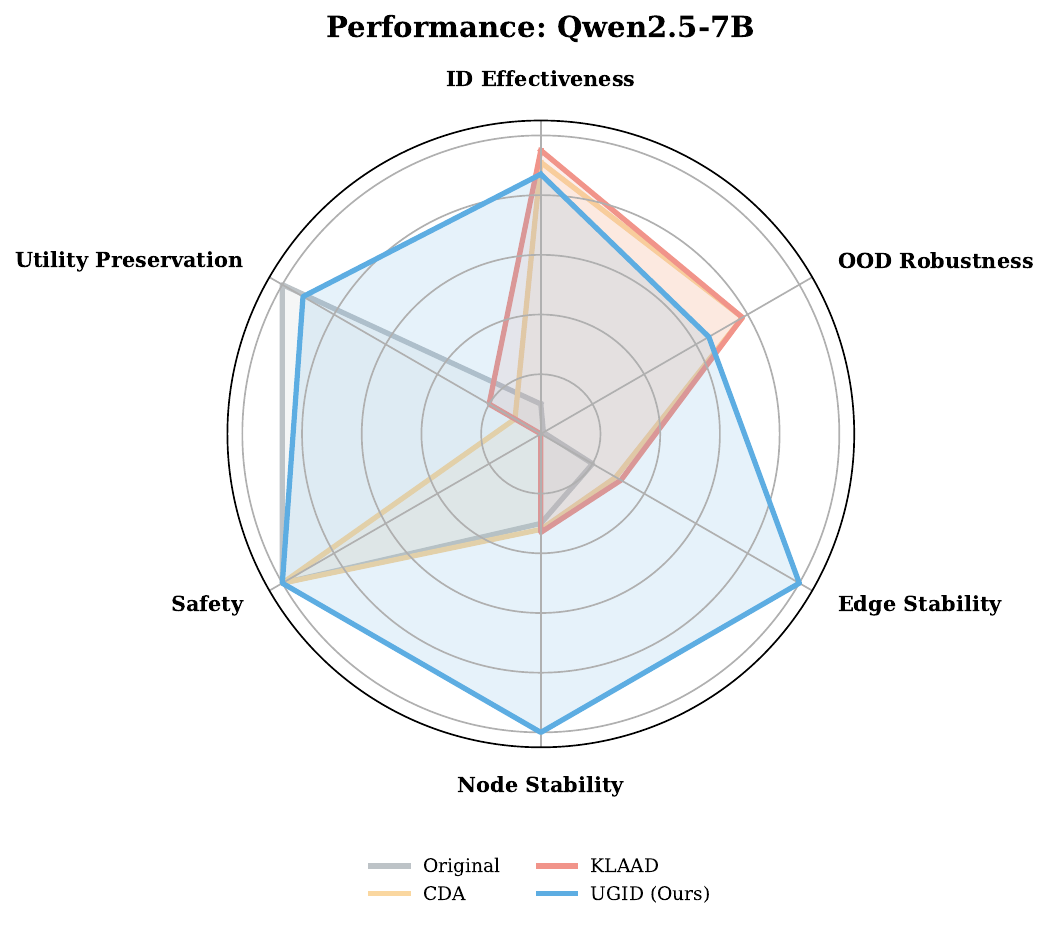}
    \caption{Qwen2.5-7B}
\end{subfigure}
\hfill
\begin{subfigure}{0.48\textwidth}
    \centering
    \includegraphics[width=\linewidth]{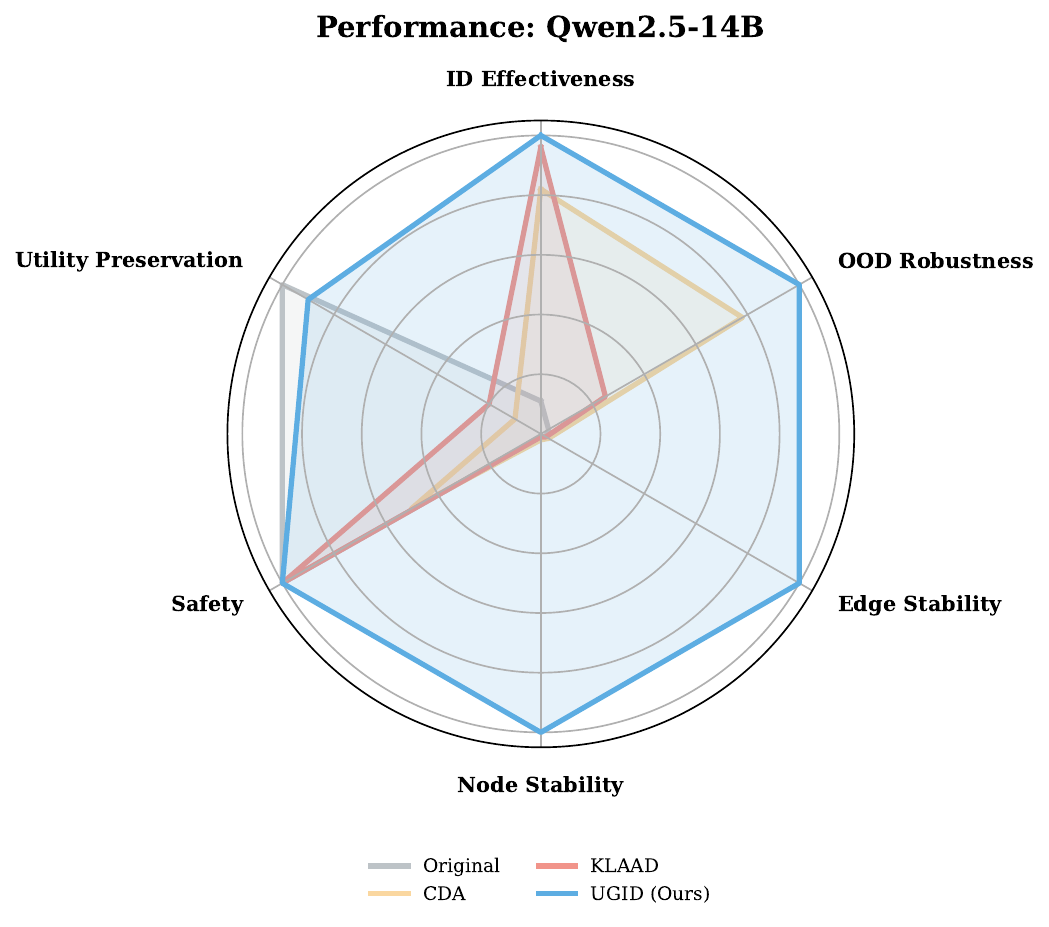}
    \caption{Qwen2.5-14B}
\end{subfigure}
\caption{\textbf{Performance Profiles across Diverse Architectures and Scales.} Across all evaluated families, \textit{UGID} consistently exhibits superior coverage (the largest blue area) compared to baselines, especially in preserving internal structural stability and safety.}
\label{fig:6}
\end{figure*}

\begin{figure}[t] 
\centering
\includegraphics[width=1.0\linewidth]{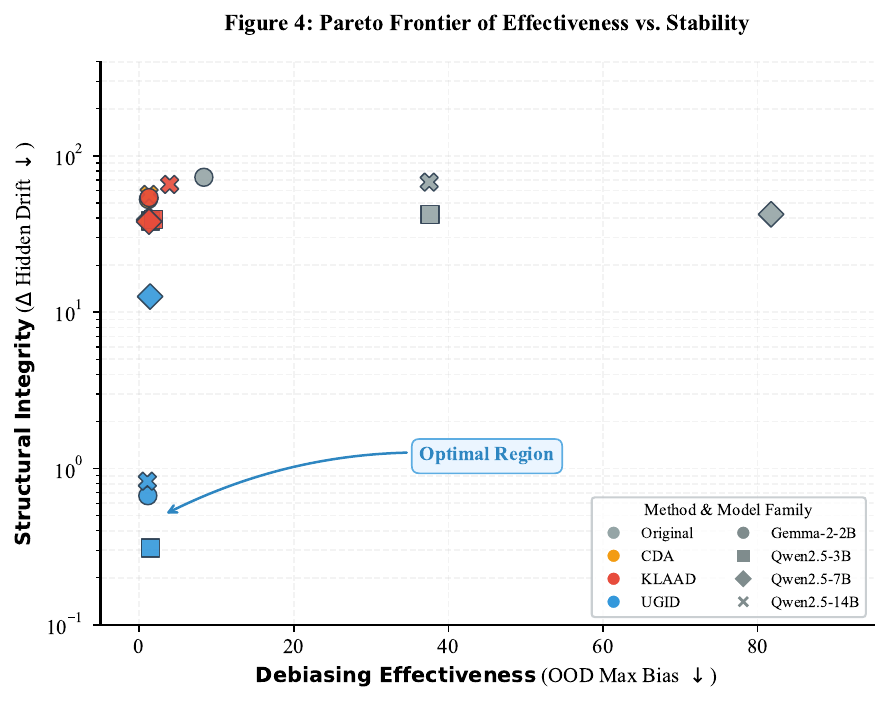}
\caption{
\textbf{Pareto Frontier of Debiasing Effectiveness vs. Structural Integrity.} 
}
\label{fig:7}
\end{figure} 

\begin{table*}[t]
\centering
\renewcommand{\arraystretch}{1.2}
\setlength{\tabcolsep}{4pt}
\resizebox{\textwidth}{!}{
\begin{tabular}{llccccccccc}
\toprule
\multirow{3}{*}{\textbf{Model Family}} 
& \multirow{3}{*}{\textbf{Method}} 
& \multicolumn{4}{c}{\textbf{Debiasing Effectiveness} ($\downarrow$ 1.0)} 
& \multicolumn{2}{c}{\textbf{Mechanism} ($\downarrow$ 0)} 
& \textbf{Safety} 
& \multicolumn{2}{c}{\textbf{Utility}} \\
\cmidrule(lr){3-6} \cmidrule(lr){7-8} \cmidrule(lr){9-9} \cmidrule(lr){10-11}
& 
& \multicolumn{2}{c}{\textbf{ID}} 
& \multicolumn{2}{c}{\textbf{OOD}} 
& \textbf{Edge} 
& \textbf{Node} 
& \textbf{Anchor} 
& \textbf{Anchor-PPL} 
& \textbf{IQ} \\
& 
& Mean & Max & Mean & Max 
& $\Delta$ Spec & $\Delta$ Hidden 
& Acc ($\uparrow$) 
& ($\downarrow$) 
& (Pass) \\
\midrule

\multirow{4}{*}{\textsc{Gemma-2-2B}} 
& \textsc{Original} 
& 6.84x & 37.45x 
& 4.48x & 8.38x 
& 0.1502 & 73.00 
& \cellcolor{paleblue}\textbf{100\%} 
& \cellcolor{paleblue}\textbf{154.79} 
& -- \\

& \textsc{CDA} 
& 1.13x & 1.21x 
& \underline{1.18x} & \underline{1.21x} 
& \underline{0.0499} & \underline{52.67} 
& \cellcolor{paleblue}\textbf{100\%} 
& 5.94 
& -- \\

& \textsc{KLAAD} 
& \underline{1.01x} & \underline{1.06x} 
& \underline{1.18x} & 1.29x 
& 0.1572 & 54.13 
& 50\% 
& 20.36 
& -- \\

& \textsc{UGID (Ours)} 
& \cellcolor{paleblue}\textbf{0.95x} 
& \cellcolor{paleblue}\textbf{1.00x} 
& \cellcolor{paleblue}\textbf{1.03x} 
& \cellcolor{paleblue}\textbf{1.13x} 
& \cellcolor{paleblue}\textbf{0.0067} 
& \cellcolor{paleblue}\textbf{0.67} 
& \cellcolor{paleblue}\textbf{100\%} 
& \underline{165.63} 
& -- \\

\midrule

\multirow{4}{*}{\textsc{Qwen2.5-3B}} 
& \textsc{Original} 
& 7.41x & 20.08x 
& 16.06x & 37.60x 
& 0.2201 & 42.04 
& \cellcolor{paleblue}\textbf{100\%} 
& \underline{114.39} 
& \checkmark \\

& \textsc{CDA} 
& 1.14x & 1.29x 
& 1.26x & \cellcolor{paleblue}\textbf{1.46x} 
& \underline{0.2133} & \underline{38.29} 
& \cellcolor{paleblue}\textbf{100\%} 
& 5.24 
& \checkmark \\

& \textsc{KLAAD} 
& \cellcolor{paleblue}\textbf{0.88x} 
& \cellcolor{paleblue}\textbf{0.88x} 
& \cellcolor{paleblue}\textbf{1.04x} 
& 1.87x 
& 0.2138 & 39.21 
& 50\% 
& 23.44 
& \checkmark \\

& \textsc{UGID (Ours)} 
& \underline{0.98x} & \underline{1.00x} 
& \underline{1.14x} & \cellcolor{paleblue}\textbf{1.46x} 
& \cellcolor{paleblue}\textbf{0.0092} 
& \cellcolor{paleblue}\textbf{0.31} 
& \cellcolor{paleblue}\textbf{100\%} 
& \cellcolor{paleblue}\textbf{101.25} 
& \checkmark \\

\midrule

\multirow{4}{*}{\textsc{Qwen2.5-7B}} 
& \textsc{Original} 
& 10.20x & 27.43x 
& 29.31x & 81.74x 
& 0.4646 & 42.29 
& \cellcolor{paleblue}\textbf{100\%} 
& \cellcolor{paleblue}\textbf{173.25} 
& \checkmark \\

& \textsc{CDA} 
& \underline{1.12x} & \underline{1.14x} 
& \underline{1.14x} & \cellcolor{paleblue}\textbf{1.29x} 
& 0.3212 & \underline{38.92} 
& \cellcolor{paleblue}\textbf{100\%} 
& 5.52 
& \checkmark \\

& \textsc{KLAAD} 
& \cellcolor{paleblue}\textbf{1.08x} 
& \cellcolor{paleblue}\textbf{1.13x} 
& \cellcolor{paleblue}\textbf{1.01x} 
& \cellcolor{paleblue}\textbf{1.29x} 
& \underline{0.2951} & 38.00 
& 0\% 
& 21.81 
& \checkmark \\

& \textsc{UGID (Ours)} 
& 1.15x & 2.24x 
& 2.14x & 1.42x 
& \cellcolor{paleblue}\textbf{0.0926} 
& \cellcolor{paleblue}\textbf{12.60} 
& \cellcolor{paleblue}\textbf{100\%} 
& \underline{188.86} 
& \checkmark \\

\midrule

\multirow{4}{*}{\textsc{Qwen2.5-14B}} 
& \textsc{Original} 
& 8.71x & 33.13x 
& 13.50x & 37.52x 
& 0.3624 & 67.96 
& \cellcolor{paleblue}\textbf{100\%} 
& \underline{170.31} 
& \checkmark \\

& \textsc{CDA} 
& 1.14x & 1.29x 
& \underline{1.16x} & \underline{1.29x} 
& \underline{0.2475} & \underline{56.67} 
& 50\% 
& 6.60 
& \checkmark \\

& \textsc{KLAAD} 
& \underline{0.98x} & \underline{1.13x} 
& 1.61x & 3.96x 
& 0.3197 & 65.67 
& \cellcolor{paleblue}\textbf{100\%} 
& 23.83 
& \checkmark \\

& \textsc{UGID (Ours)} 
& \cellcolor{paleblue}\textbf{0.94x} 
& \cellcolor{paleblue}\textbf{1.00x} 
& \cellcolor{paleblue}\textbf{1.01x} 
& \cellcolor{paleblue}\textbf{1.07x} 
& \cellcolor{paleblue}\textbf{0.0075} 
& \cellcolor{paleblue}\textbf{0.83} 
& \cellcolor{paleblue}\textbf{100\%} 
& \cellcolor{paleblue}\textbf{153.71} 
& \checkmark \\

\bottomrule
\end{tabular}
}
\caption{\textbf{Results on Diverse Model Families.} \colorbox{paleblue}{Paleblue} and \underline{underline} indicate the best and second-best results within each model family (excluding Original baseline).}
\label{tab:7}
\vspace{-5mm}
\end{table*}
\begin{figure}[t]
\centering
\includegraphics[width=\linewidth]{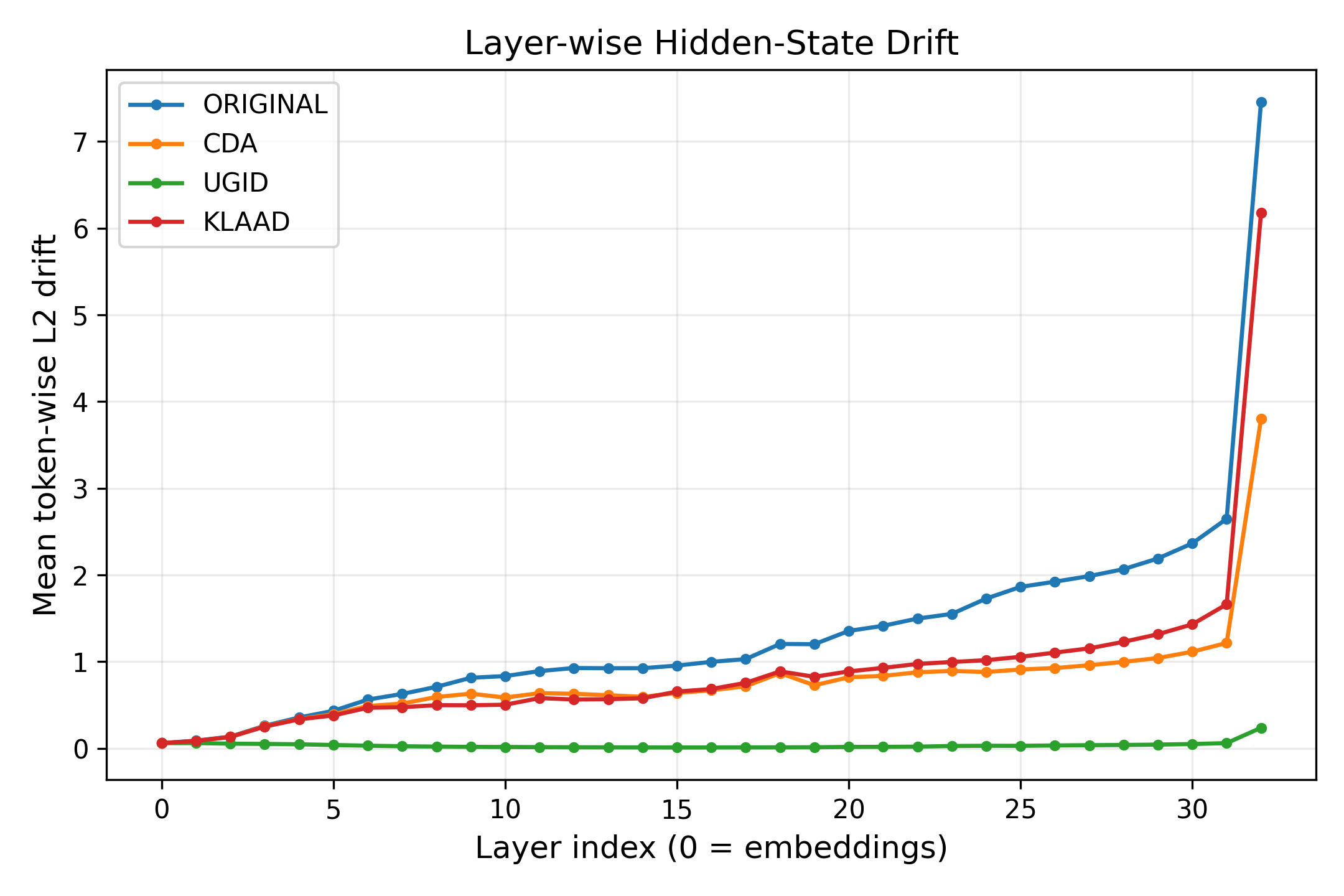}
\caption{
\textbf{Layer-wise Hidden-State Drift under Counterfactual Pronoun Swaps.}
}
\label{fig:8}
\end{figure}

\begin{figure}[t]
    \centering
    \includegraphics[width=\linewidth]{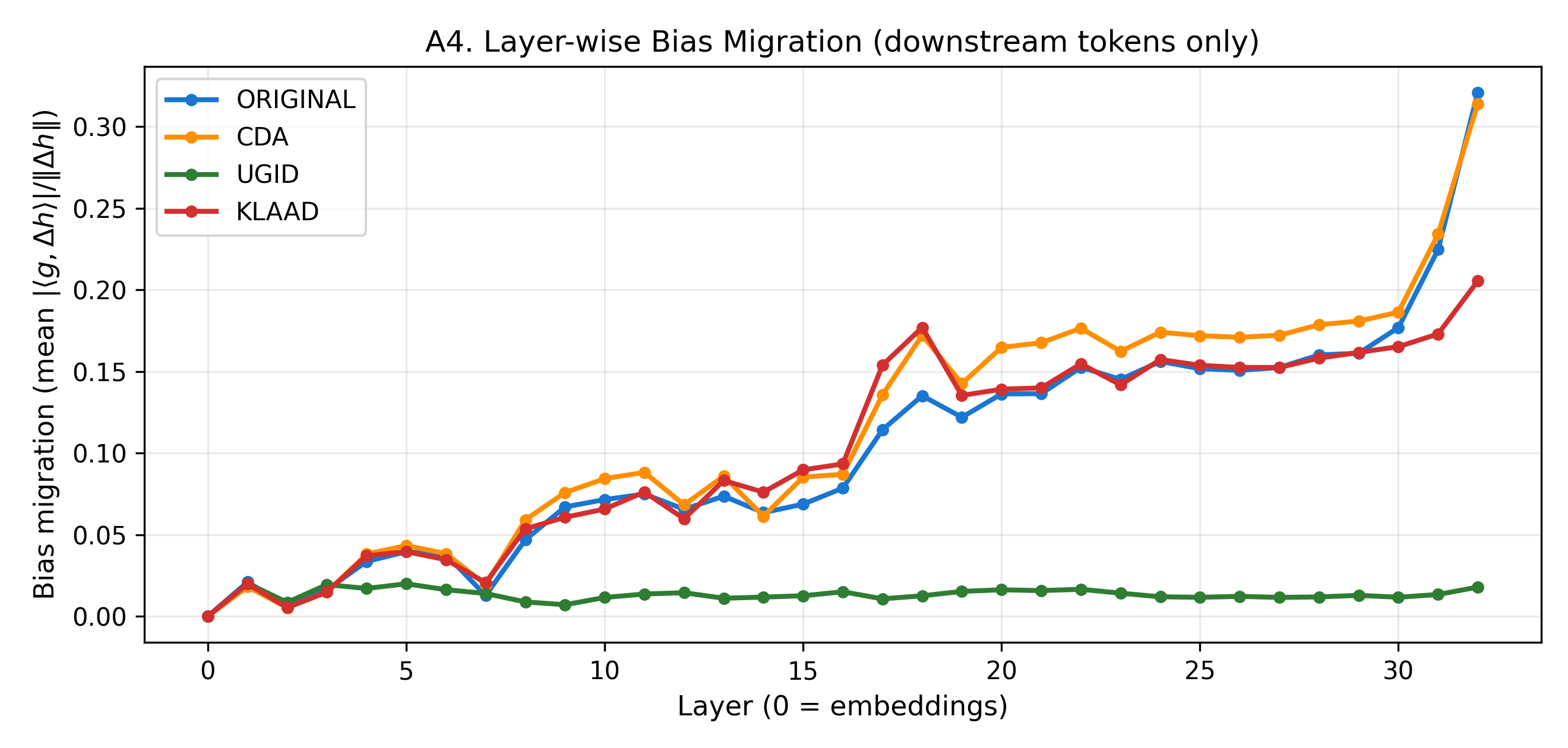}
    \vspace{-1em}
    \caption{\textbf{Layer-wise bias migration.}}
    \label{fig:9}
    \vspace{-0.8em}
\end{figure}

\begin{figure}[t]
    \centering
    \includegraphics[width=\linewidth]{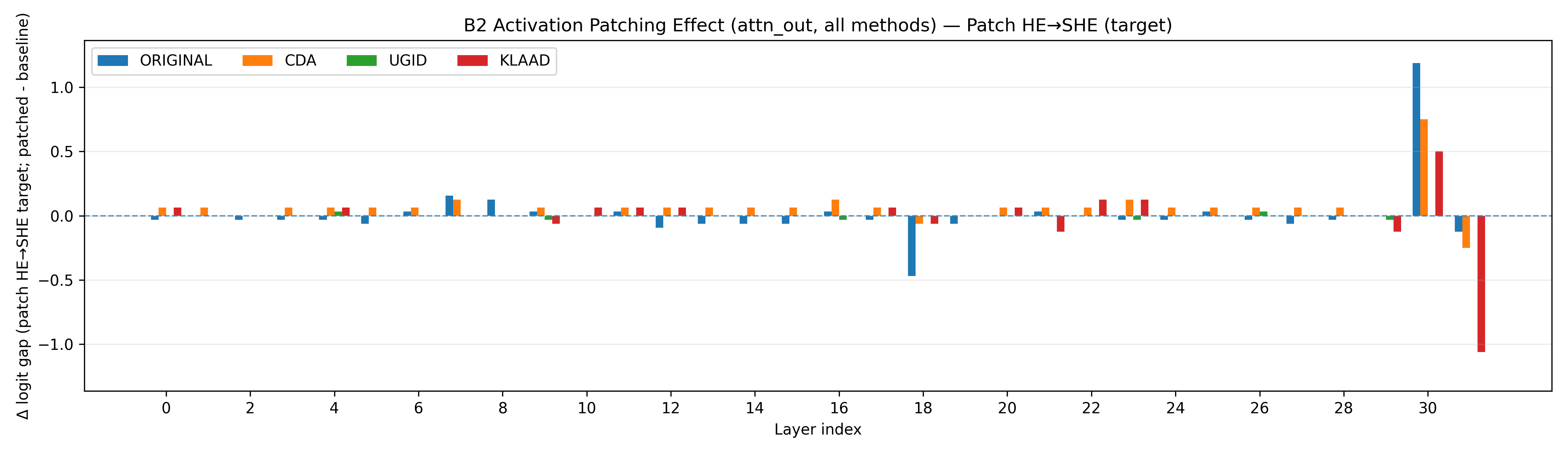}
    \vspace{-1em}
    \caption{\textbf{Activation patching / causal tracing results (He to She).}}
    \label{fig:10}
    \vspace{-0.8em}
\end{figure}

\begin{figure}[t]
    \centering
    \includegraphics[width=\linewidth]{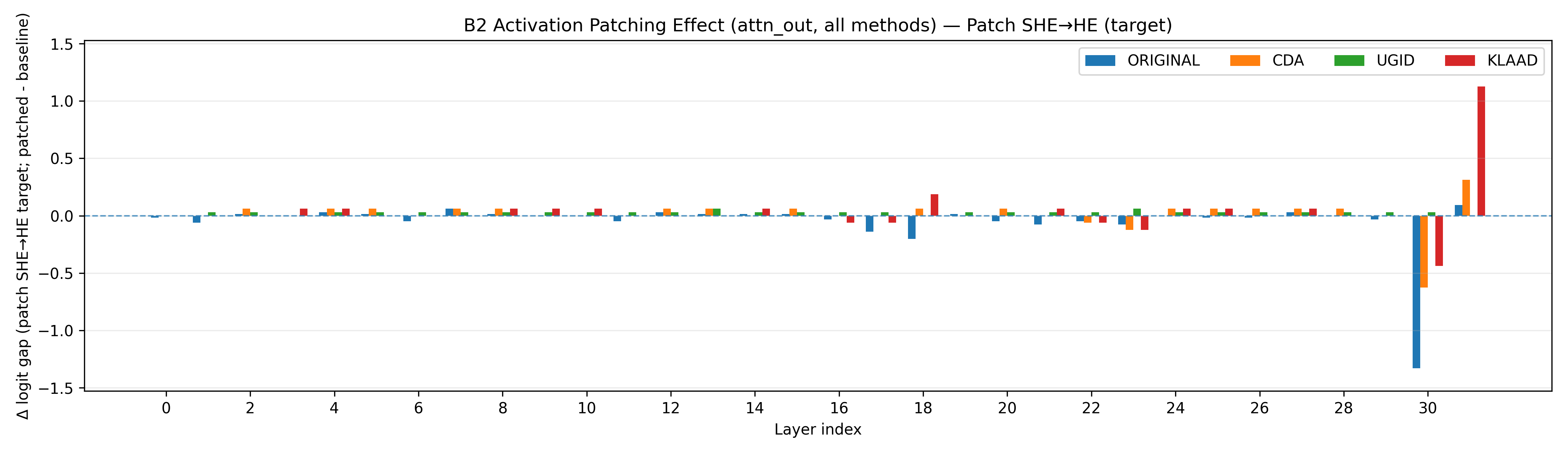}
    \vspace{-1em}
    \caption{\textbf{Activation patching / causal tracing results (She to He).}}
    \label{fig:11}
    \vspace{-0.8em}
\end{figure}

\begin{figure*}[t]
  \centering
  \setlength{\tabcolsep}{3pt}
  \begin{tabular}{cccc}
    \includegraphics[width=0.5\textwidth]{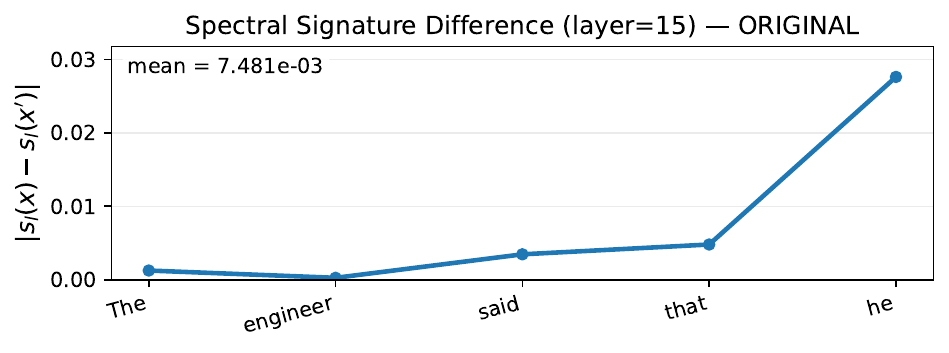} &
    \includegraphics[width=0.5\textwidth]{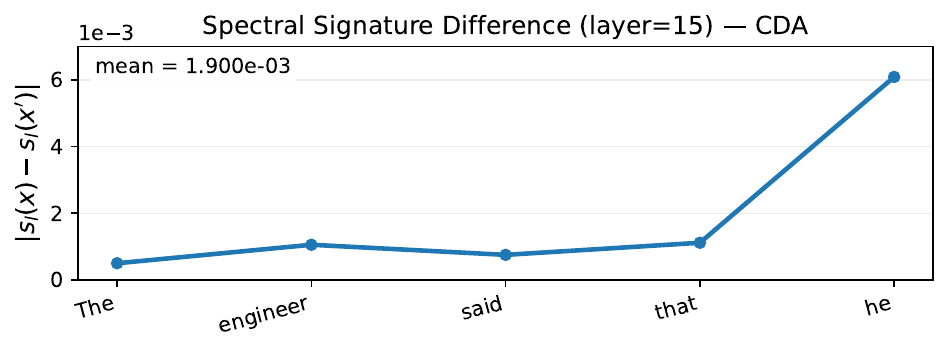} \\
  \end{tabular}
  \caption{
  \textbf{Spectral Signature Difference (layer=15).}
  Token-wise $|s_l(x)-s_l(x')|$ for \emph{he/she} counterfactuals (Original and CDA).
  }
  \label{fig:12}
\end{figure*}

\begin{figure*}[t]
  \centering
  \setlength{\tabcolsep}{3pt}
  \begin{tabular}{cccc}
    \includegraphics[width=0.5\textwidth]{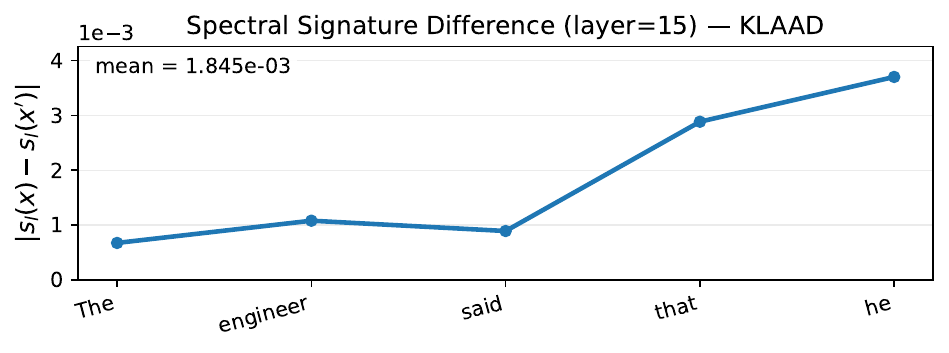} &
    \includegraphics[width=0.5\textwidth]{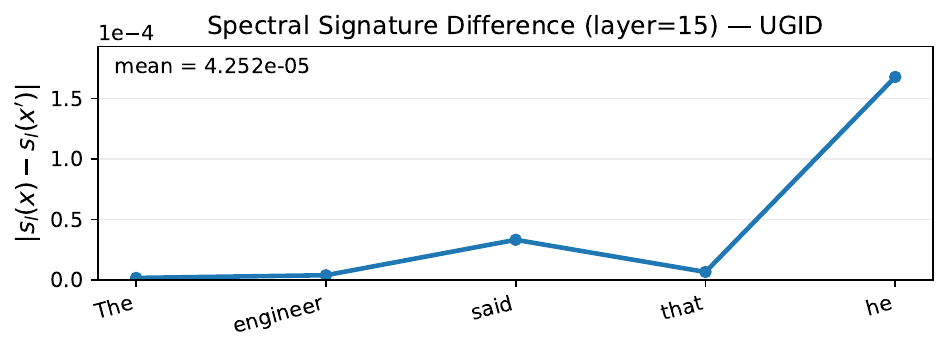} \\
  \end{tabular}
  \caption{
  \textbf{Spectral Signature Difference (layer=15).}
  Token-wise $|s_l(x)-s_l(x')|$ for \emph{he/she} counterfactuals (KLAAD and UGID). 
  }
  \label{fig:13}
\end{figure*}

\end{document}